\documentclass{article}

\usepackage{microtype}
\usepackage{graphicx}
\usepackage{booktabs} 

\usepackage{hyperref}
\usepackage{tikz}
\usepackage{amsmath}
\usepackage{amssymb}
\usepackage{xspace}
\usepackage{comment}
\usepackage{caption}
\usepackage{subcaption}
\usepackage{soul}
\usepackage{multirow}
\newcommand{\codename}{TeraPipe\xspace}

\DeclareMathOperator*{\argmin}{argmin}
\newif\ifcomments
\commentstrue 
\ifcomments
    \providecommand{\zhuohan}[1]{{\protect\color{blue}{[Zhuohan: #1]}}}
    \providecommand{\danyang}[1]{{\protect\color{purple}{ [Danyang: #1]}}}
    \providecommand{\siyuan}[1]{{\protect\color{red}{[Siyuan: #1]}}}
    \providecommand{\hao}[1]{{\protect\color{green}{[Hao: #1]}}}
    \providecommand{\jeff}[1]{{\protect\color{orange}{[Jeff: #1]}}}
\else
    \providecommand{\zhuohan}[1]{}
    \providecommand{\danyang}[1]{}
    \providecommand{\siyuan}[1]{}
    \providecommand{\hao}[1]{}
    \providecommand{\jeff}[1]{}
\fi

\usepackage[accepted]{icml2021}

\icmltitlerunning{\codename: Token-Level Pipeline Parallelism for Training Large-Scale Language Models}

\begin{document}

\twocolumn[
\icmltitle{\codename: Token-Level Pipeline Parallelism for Training \\ Large-Scale Language Models}



\icmlsetsymbol{equal}{*}

\begin{icmlauthorlist}
\icmlauthor{Zhuohan Li}{berkeley}
\icmlauthor{Siyuan Zhuang}{berkeley}
\icmlauthor{Shiyuan Guo}{berkeley}
\icmlauthor{Danyang Zhuo}{duke}
\icmlauthor{Hao Zhang}{berkeley}
\icmlauthor{Dawn Song}{berkeley}
\icmlauthor{Ion Stoica}{berkeley}
\end{icmlauthorlist}

\icmlaffiliation{berkeley}{UC Berkeley}
\icmlaffiliation{duke}{Duke University}

\icmlcorrespondingauthor{Zhuohan Li}{zhuohan@cs.berkeley.edu}

\icmlkeywords{Machine Learning, ICML}

\vskip 0.3in
]



\printAffiliationsAndNotice{} 

\begin{abstract}
Model parallelism has become a necessity for training modern large-scale deep language models. In this work, we identify a new and orthogonal dimension from existing model parallel approaches: it is possible to perform pipeline parallelism within a single training sequence for Transformer-based language models thanks to its autoregressive property. This enables a more fine-grained pipeline compared with previous work. With this key idea, we design \codename, a high-performance token-level pipeline parallel algorithm for synchronous model-parallel training of Transformer-based language models. We develop a novel dynamic programming-based algorithm to calculate the optimal pipelining execution scheme given a specific model and cluster configuration. We show that \codename can speed up the training by 5.0x for the largest GPT-3 model with 175 billion parameters on an AWS cluster with 48 p3.16xlarge instances compared with state-of-the-art model-parallel methods. The code for reproduction can be found at \url{https://github.com/zhuohan123/terapipe}
\end{abstract}

\begin{figure*}[t]
    \centering
    \begin{subfigure}[t]{.24\textwidth}
        \centering
        \includegraphics[ height=4.2cm]{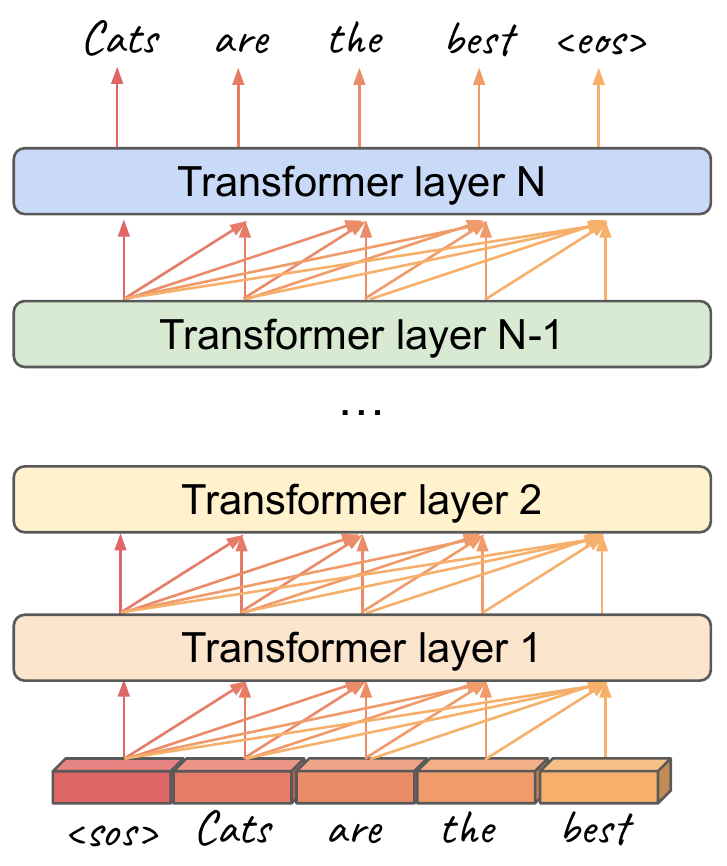}
        \caption{Transformer-based LM}
        \label{fig:transformer}
        \end{subfigure}
    \begin{subfigure}[t]{.24\textwidth}
        \centering
        \includegraphics[height=4.2cm]{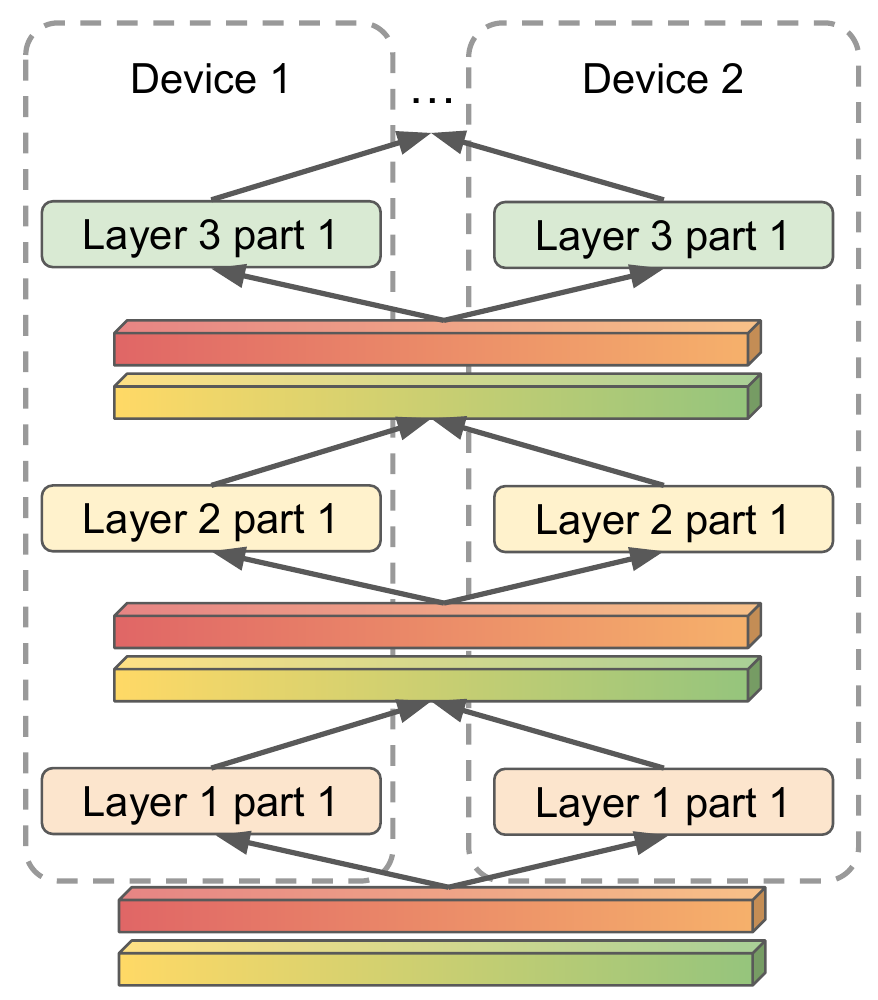}
        \caption{Operation partitioning (Megatron-LM)}
        \label{fig:megatron}
    \end{subfigure}
    \begin{subfigure}[t]{.24\textwidth}
        \centering
        \includegraphics[height=4.2cm]{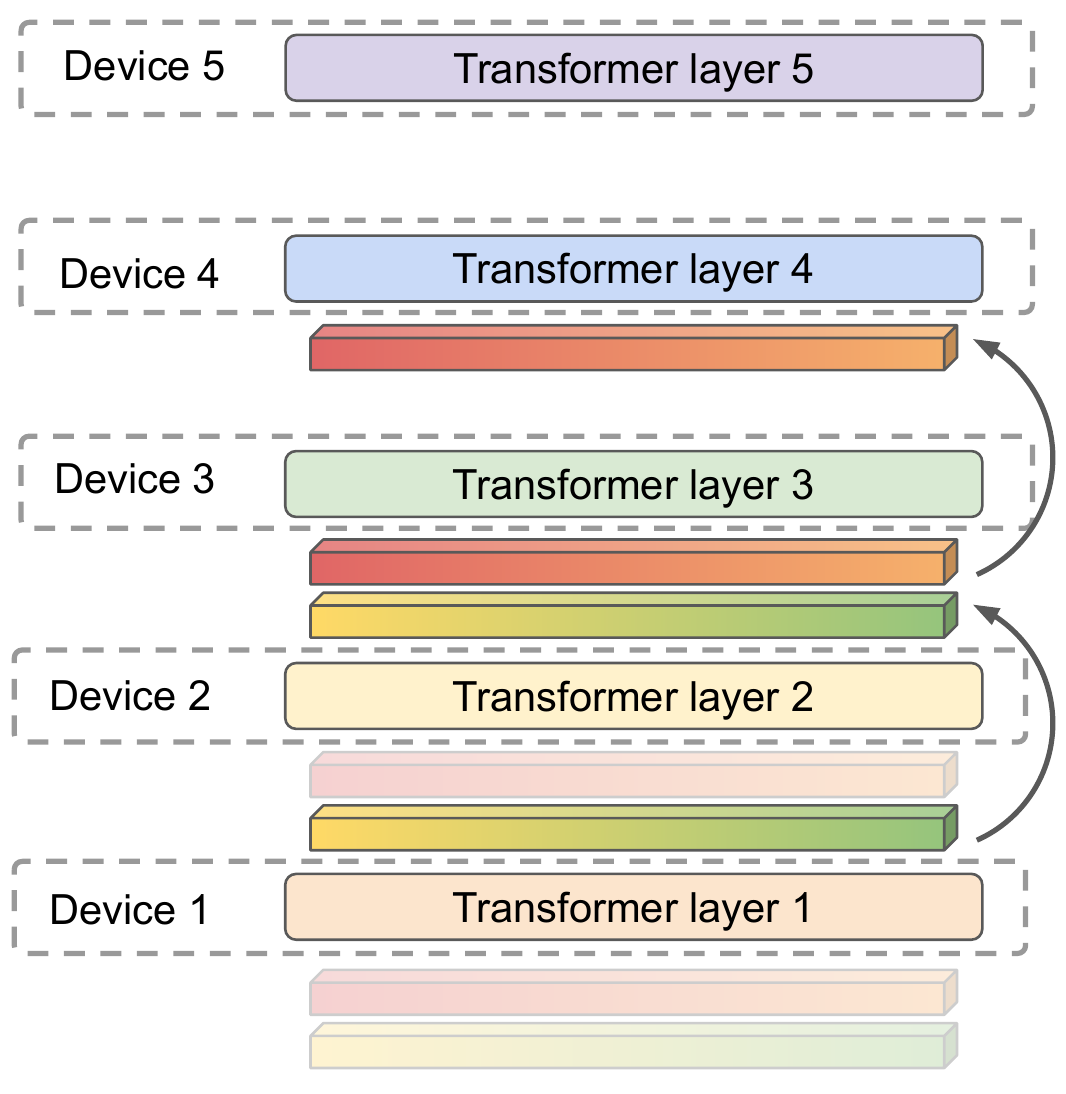}
        \caption{Microbatch-based pipeline parallelism (GPipe)}
        \label{fig:gpipe}
    \end{subfigure}
    \begin{subfigure}[t]{.24\textwidth}
        \centering
        \includegraphics[height=4.2cm]{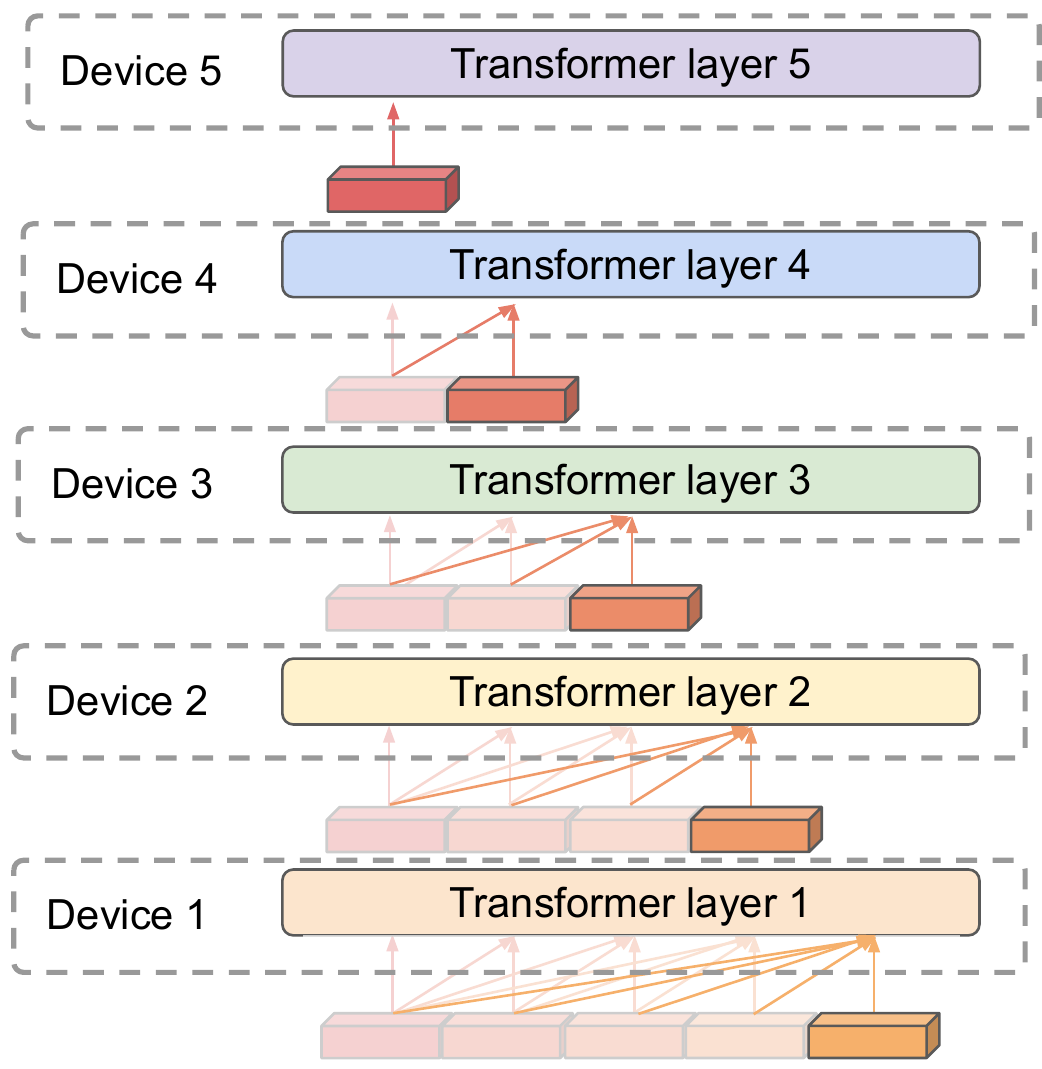}
        \caption{Token-based pipeline parallelism (\codename)}
        \label{fig:terapipe}
    \end{subfigure}
    \caption{Different approaches of model parallel training of Transformer-based LMs. (a) shows a standard multi-layer Transformer LM. In each layer, each position only takes only its previous positions as input. (b) shows operation partitioning \citep{shoeybi2019megatron}. An allreduce operation is required to synchronize the results of each layer. (c) shows microbatch-based pipeline parallelism \citep{huang2019gpipe}, which allows different microbatches (red and green bars) to be executed on different layers of the DNN in parallel. (d) show \codename (our work), which pipelines along the token dimension.}
        \vspace{-4mm}
    \label{fig:my_label}
\end{figure*}

\section{Introduction}

Transformer-based language models (LMs) have revolutionized the area of natural language processing (NLP) by achieving state-of-the-art results for many NLP tasks, including text classification, question answering, and text generation \citep{brown2020language, radford2019language}.
The accuracy of a Transformer-based LM grows substantially with its model size, attributing to the fact that they can be \emph{unsupervisedly} trained on almost \emph{unlimited} text data. Today, a large LM, such as GPT-3 \citep{brown2020language}, can have more than 175B parameters, which amounts to 350\,GB, assuming 16-bit floating-point numbers. 
This significantly exceeds the memory capacity of existing hardware accelerators, such as GPUs and TPUs, which makes model-parallel training a necessity, i.e., partitioning the model on multiple devices during the training process.

Because of the demands for efficient LM training, many researchers and industry practitioners have proposed different ways for model parallel training. One approach is to partition the weight matrices and dispatch smaller matrix operations to parallel devices \citep[Figure~\ref{fig:megatron};][]{shoeybi2019megatron,shazeer2018mesh}. Another approach is to split a batch of training data into many microbatches and then evenly pipeline the layer computations across different microbatches and devices \citep[Figure~\ref{fig:gpipe};][]{huang2019gpipe}. 
Unfortunately, these approaches either introduce excessive communication overheads between compute devices, or lead to reduced efficiency due to pipeline ``bubbles'' (i.e. device idle time, see Section~\ref{sec:related-work} and \ref{sec:main-technique} for details). 

Our key observation in this paper is that Transformer-based language models have a key property: the computation of a given input token only depends on previous tokens, but not on future tokens. This lack of dependency on future tokens provides new opportunities for pipeline parallel training.\footnote{In this paper, we focus on unidirectional autoregressive language models (e.g., GPT~\citep{radford2019language, brown2020language}) but not bidirectional models like masked language models (e.g., BERT~\citep{devlin2018bert}).} In particular, it allows us to create a fine-grained pipeline within a single training sequence for Transformer-based LMs, by parallelizing the computation of the current token on the current layer with the computation of the previous token on the next layer of the model. For example, in Figure~\ref{fig:terapipe}, we can pipeline the execution across all 5 devices within a single input sequence.
Similar to other synchronous model parallel training methods, e.g., Gpipe~\cite{huang2019gpipe}, Megatron-LM~\cite{shoeybi2019megatron}, we do not change the underlying optimization algorithm, so the resulting model has exactly the same accuracy.

However, leveraging the token dimension for efficient model parallel training raises several challenges.
First, if the partitioning along the token dimension is too fine-grained, it leads to under-utilization on devices that require large blocks of data for efficient processing (e.g., GPU).
Second, since each token position in the sequence depends on all previous tokens, different positions in a transformer layer exhibit uneven computation loads.
This means that uniformly partitioning along the token dimension might cause uneven load across devices, and degenerate the training efficiency.

To this end, we design and implement \emph{\codename}, a high-performance synchronous model parallel training approach for large-scale Transformer-based language models, which exploits the token dimension to pipeline the computation across devices. \codename uses a small number of simple workloads to derive a performance model and then uses a novel dynamic programming algorithm to compute the optimal partitioning of the token dimension for the pipeline. \codename is orthogonal to previous model-parallel training methods, so it can be used together with these methods to further improve the training performance. 
Our evaluation shows that for the largest GPT-3 model with 175 billion parameters,~\codename achieves a 5.0x speedup improvement over the state-of-the-art synchronous model-parallel training methods on an AWS cluster consisting of 48 p3.16xlarge instances.

Our paper makes the following contributions:
\begin{itemize}
    \item We propose a new dimension, token dimension, for pipeline-parallel training of Transformer-based LMs.
    \item We develop a dynamic programming algorithm to compute a partition along the token dimension to maximize pipeline parallelism.
    \item We implement \codename and show that we can increase the synchronous training throughput of the largest GPT-3 model (with 175 billion parameters) by 5.0x over the previous state-of-the-art model-parallel methods.
\end{itemize}

\section{Related Work}
\label{sec:related-work}

\textbf{Data parallelism} scales ML training by partitioning training data onto distributed devices 
\cite{parallelsgd2010, Krizhevsky2014OneWT, goyal2017accurate, rajbhandari2019zero}. Each device holds a model replica, works on an independent data partition, and synchronizes the updates via \emph{allreduce}  \citep{Krizhevsky2014OneWT} or a parameter server \citep{li2014scaling}. Data parallelism alone is not enough to train large-scale DNNs due to two main reasons: (1) every device has to have enough memory to store the model and the gradients generated during the training process; (2) communication can be a performance bottleneck to synchronize model parameters.

\textbf{Model parallelism} allows for training models larger than the memory capacity of a single device, by partitioning the model (e.g., layers) into disjoint parts and executing each on a dedicated device. Existing model parallel training approaches can be roughly categorized as: \emph{operation partitioning} and \emph{pipeline parallelism}.

\textbf{Operation partitioning.} One way to split the model is to partition and parallelize computational operations across multiple devices. For example, the computation of matrix multiplications (matmul) $XAB$ can be spitted across multiple devices by partitioning $A$ and $B$ along its rows and columns, respectively.
\[
XAB = X\cdot\begin{bmatrix} A_1 & A_2 \end{bmatrix} \cdot \begin{bmatrix} 
    B_1 \\
    B_2
\end{bmatrix} = X A_1 B_1 + X A_2 B_2.
\]
This means we can have one device calculate $X A_1 B_1$ and another device calculate $X A_2 B_2$ in parallel. After that, cross-device communication is needed to compute the sum of these two parts.

Many existing works \citep{jia2018flexflow,jia2019beyond, wang2019supporting, shazeer2018mesh} study how to optimize the partitioning schemes for different operations to maximize throughput and minimize communication overheads, among which, Megatron-LM \citep[Figure~\ref{fig:megatron};][]{shoeybi2019megatron} designs partitioning schemes specifically for large-scale Transformers.
However, due to the excessive communication required to collect partial results after each layer, 
it 
is not efficient when the bandwidth between devices is limited \citep{shoeybi2019megatron}. Flexflow \citep{jia2018flexflow} proposes a framework to find the optimal operation partitioning, but it cannot model the new dimension proposed in our work.

\textbf{Pipeline parallelism} partitions a DNN into layers and put different layers onto different devices
\citep[Figure~\ref{fig:gpipe};][]{petrowski1993pipelinedbp}. 
Each device computes the input on a given layer and sends the result to the next device.
Pipeline parallelism significantly reduces communication between devices, because only devices holding neighboring layers need to communicate and they only need to communicate the activations on a particular layer. 

\begin{figure}[t]
    \centering
    \begin{subfigure}[t]{.47\textwidth}
        \centering
        \includegraphics[width=\textwidth]{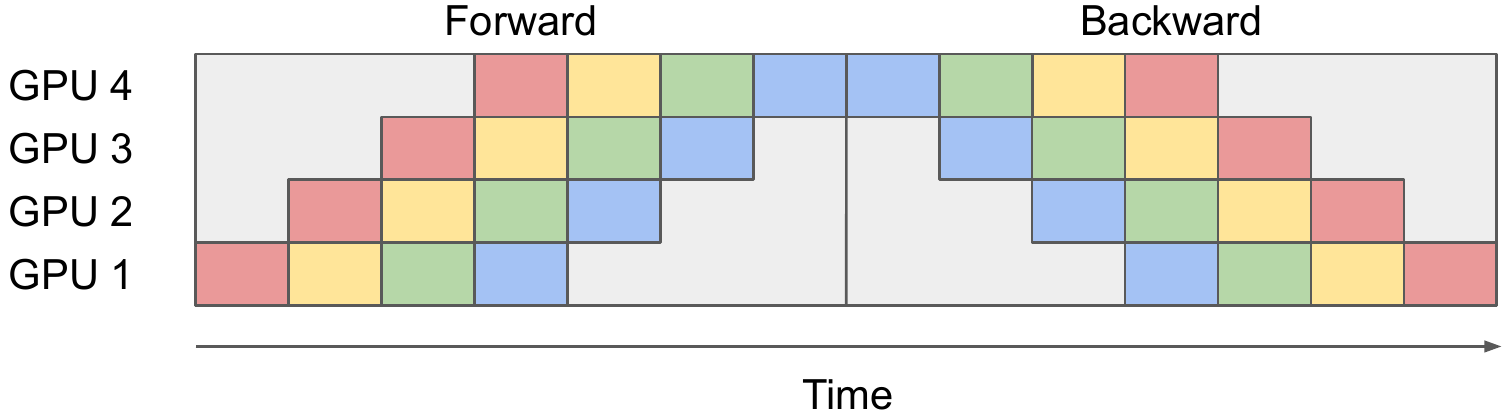}
        \caption{Microbatch-based pipeline parallelism}
        \label{fig:gpipe-big-timeline}
    \end{subfigure}
    \begin{subfigure}[t]{.47\textwidth}
        \centering
        \includegraphics[width=\textwidth]{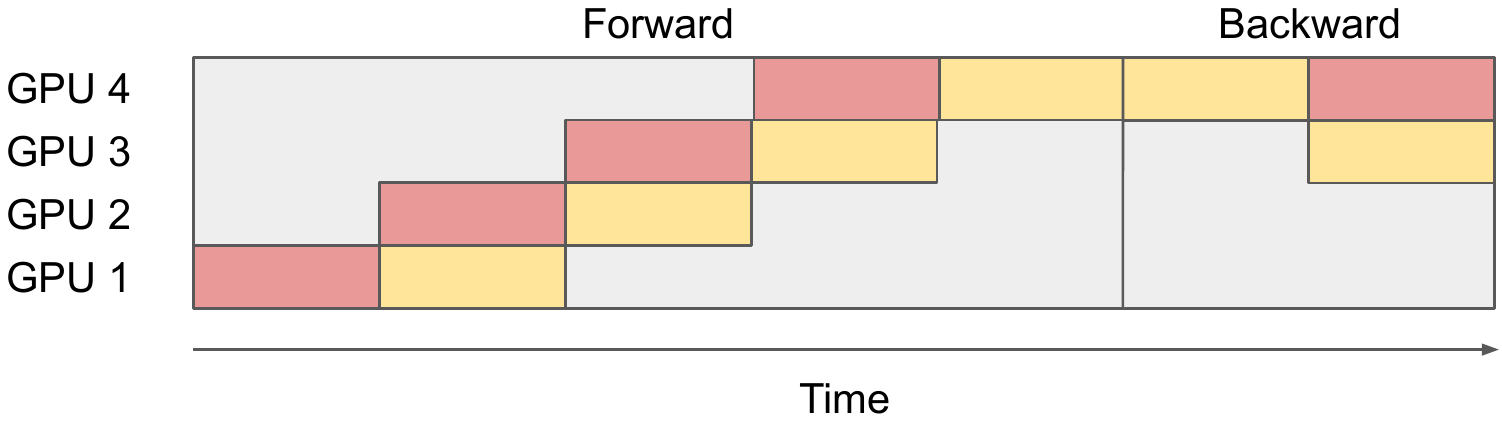}
        \caption{Microbatch-based pipeline parallelism with small batch size}
        \label{fig:gpipe-timeline}
    \end{subfigure}
    \begin{subfigure}[t]{.47\textwidth}
        \centering
        \includegraphics[width=\textwidth]{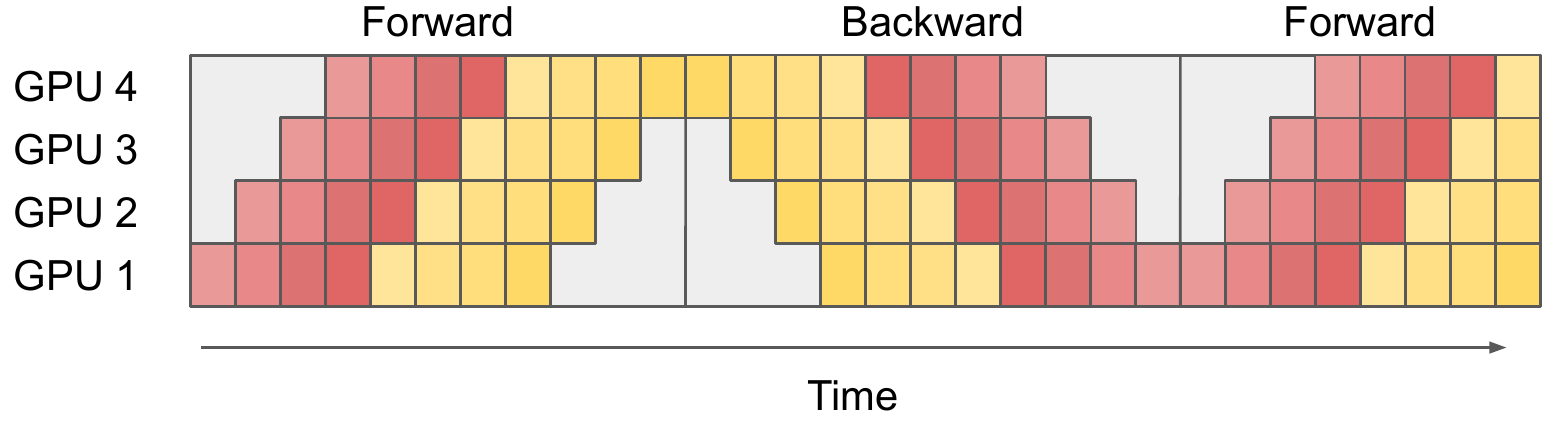}
        \caption{\codename}
        \label{fig:terapipe-timeline}
    \end{subfigure}
        \vspace{-2mm}
    \caption{\small Execution timeline for different pipelining methods. Grey blocks indicate GPUs idle time (a.k.a. pipeline bubbles). (a) Microbatch-based pipeline parallelism (e.g. GPipe). Each color corresponds to a microbatch. (b) Microbatch-based pipeline parallelism with longer sequence (hence smaller minibatch size due to fixed GPU memory). Pipeline bubbles significantly increase. (c) \codename. Pipeline bubbles are substantially reduced because of the improved pipelining granularity.}
        \vspace{-4mm}

    \label{fig:timeline}
\end{figure}

Previous pipeline parallel training methods are based on \emph{microbatch} pipelining, e.g., GPipe~\cite{huang2019gpipe}. This means the computation for a given microbatch in a minibatch on a layer can run in parallel with the next microbatch in the same minibatch on the previous layer.
However, microbatch-based pipeline parallelism still cannot achieve high efficiency due to its pipeline bubbles. This is because the start of the forward propagation on a minibatch requires the backward propagation of the previous minibatch to complete (Figure \ref{fig:gpipe-big-timeline}). This problem becomes more severe when model sizes increase (see Section \ref{sec:main-technique}).
\citet{harlap2018pipedream} propose using an asynchronous training algorithm to mitigate the effect of pipeline bubbles in microbach-based pipeline parallel training, but asynchronous training introduces uncertainty in model accuracy and is thus not widely adopted for training DNNs.

\textbf{Wavefront parallelism} is a variant of pipeline parallelism, broadly applied in shared-memory multiprocessors \citep{sinharoy1994wavefront, manjikian1996scheduling}. In deep learning, it has been used to accelerate the computation of multi-layer RNNs on a single GPU \citep{appleyard2016optimizing}, where different input positions of different layers can execute in parallel in a wavefront fashion to maximize the utilization of the GPU. However, wavefront parallelism cannot accelerate the execution of Transformers because there is no dependency between different input positions within a single Transformer layer to begin with. In addition, wavefront parallelism uses fine-grained per-word pipelining due to the temporal data dependency in RNNs, while too fine-grained pipelining in \codename would lead to inferior pipeline efficiency (see Section~\ref{sec:main-technique} and \ref{sec:dp}). 

\section{Method}

In this section, we briefly introduce language modeling and Transformers. Based on their structures, we identify new opportunities for performing pipelining along the input sequence (which we will notate as the \emph{token dimension} in the rest of the paper).
With that, we derive the optimal slicing scheme over the token dimension to maximize pipeline efficiency using a dynamic programming algorithm. Finally, we show how to combine our new method with existing parallel training techniques.

\subsection{Language Modeling and Transformers}

The task of language modeling is usually framed as unsupervised distribution estimation of a text corpus $\mathcal{X}$, where each example $x \sim \mathcal{X}$ is a variable length sequence of tokens $(x_1, x_2, \ldots, x_L).$ Since language has a natural sequential ordering, it is common to factorize the joint probability over the tokens as the product of conditional probabilities \citep[a.k.a. autoregressive decomposition;][]{bengio2003neural}:
\begin{equation}
\small
    \label{eq:conditional_prob}
    P(x) = \prod_{t=1}^L P(x_t|x_1, \ldots, x_{t-1}).
\end{equation}
Transformer~\citep{vaswani2017attention} is the state-of-the-art architecture for modeling these conditional probabilities.
As visualized in Figure~\ref{fig:transformer}, a Transformer-based LM $F$ takes the sequence $(\left<\mathit{sos}\right>, x_1, \ldots, x_{L-1})$ as input, where $\left<\mathit{sos}\right>$ represents the start of a sentence, and outputs a probability distributions $p_t$ at each position $t$ that models the conditional probability $P(x_t|x_1, \ldots, x_{t-1})$ as in Eq.~\ref{eq:conditional_prob}. In practice, $F$ is stacked with many Transformer layers $F = f_N \circ f_{N-1} \circ \cdots \circ f_1$ \cite{vaswani2017attention, radford2019language}:
$f_1$ takes the embedding of the original sequence as input, while $f_i$ ($i>1$) takes the output of $f_{i-1}$ as input.
The main components of a Transformer layer $f$ contain a \emph{self-attention layer} and a \emph{position-wise feed-forward network layer}:
\begin{align}
\small
    &\text{SelfAtt}(h_t; h_1, \ldots, h_{t-1}) = \sum_{s=1}^{t} \alpha_{ts}\cdot(W_V h_{s}), \nonumber \\
    &\text{where } \alpha_{ts} = \text{softmax}\left(\frac{(W_Qh_t)^\top(W_Kh_s)}{ \sqrt{H}}\right); \label{eq:selfatt}\\
    &\text{FFN}(h_t) = W_2 \sigma(W_1 h_t + b_1) + b_2. \label{eq:ffn}
\end{align}
$h_1, \ldots, h_L \in \mathbb{R}^{H}$ are hidden states correspond to each position of the input sequence, $W$ and $b$ are learnable parameters, and $\sigma$ is the nonlinear activation function. An important note here: for each $h_t$, Eq.~\ref{eq:selfatt} takes only the hidden states before position $t$ as inputs and Eq.~\ref{eq:ffn} only takes $h_t$ as input.

The operation and data dependency in Transformers make it more amenable to parallelization on GPUs/TPUs compared to RNNs \citep{vaswani2017attention}.  Therefore, Transformers have been scaled to enormous datasets and achieved state-of-the-art performance on a wide range of NLP tasks \citep{vaswani2017attention,devlin2018bert,radford2019language,yang2019xlnet,brown2020language,liu2019roberta}. 
Recently, people show that the accuracy of LMs can consistently improve with increasing model sizes \citep{radford2019language, yang2019xlnet}. While the growing model size greatly exceeds the memory capacity of a single GPU \citep{brown2020language}, model parallelism becomes a necessity for training large-scale LMs \citep{shoeybi2019megatron}.

\subsection{Pipeline Parallelism Within a Sequence}
\label{sec:main-technique}

In this subsection, we expose the limitations of existing pipelining parallelism approaches, and develop the proposed new pipelining method for Transformer-based LMs. 

Typically, to perform pipeline parallelism, a Transformer model $F$ is partitioned into multiple cells $c_1, \ldots, c_K$. Each cell $c_k$ consists of a set of consecutive Transformer layers $f_j \circ \cdots \circ f_{i+1} \circ f_i$ so that $F = c_K\circ \cdots \circ c_2 \circ c_1$. Each $c_k$ is placed and executed on the $k$-th device (e.g. GPU). The output of cell $c_k$ is sent to cell $c_{k+1}$ during forward propagation, and the backward states computed on cell $c_{k+1}$ is sent to cell $c_k$ during backward propagation. 
Since each layer $f$ exhibits the same structure, the entire LM can be uniformly partitioned: each cell possesses the same number of layers hence the same amount of computation workload, to reach optimal pipeline efficiency (see Figure~\ref{fig:timeline}). 

However, previous pipelining methods \citep{huang2019gpipe, harlap2018pipedream} do not perform well on large Transformer-based LMs due to the growing model size. 
Consider a minibatch of size $B$. The input to a Transformer layer $f$ is a 3-dimensional tensor $(h^{(1)}, h^{(2)}, \ldots, h^{(B)})$ of size $(B, L, H),$ where $L$ is the sequence length and $H$ is the hidden state size. To improve accuracy, large LMs are often configured to have a large $L$ to capture longer-term dependency in language sequences \citep{tay2020long, zaheer2020big}.
To fit the model into a GPU, the minibatch size $B$ has to decrease accordingly. The pipeline bubbles become larger (Figure \ref{fig:gpipe-timeline}) because fewer input sequences can be processed in parallel. 

In this work, we make a key observation: for Transformer-based LMs, with appropriate scheduling, the \emph{token dimension} $L$ can be pipelined for parallel training; and this pipelining dimension is complementary to other model parallelism approaches. 
Precisely, for an input hidden state sequence $(h_1, h_2, \ldots, h_{L})$, the computation of a self-attention layer $\text{SelfAtt}(h_t)$ only depends on the hidden states of previous positions $(h_1, \ldots, h_{t-1})$, and the computation of a feed-forward layer $\text{FFN}(h_t)$ only depends on $h_t$ itself. These offer a new opportunity for pipelining: 
the computation of layer $f_i$ at step $t$ can commence once the hidden states of previous steps ($<t$) at $f_{i-1}$ are ready, which, also, can be parallelized with the computation of latter steps at $f_{i-1}$, illustrated in Figure~\ref{fig:terapipe}.
This property enables us to perform pipeline parallelism within a single input sequence. Specifically, we can split an input sequence $x_1, \ldots, x_L$ into $s_1, \ldots, s_M$, where each subsequence $s_i$ consists of tokens $(x_l, x_{l+1}, \ldots, x_{r}).$ The computation of $c_1, \ldots, c_K$ over $s_1, \ldots, s_M$ can be pipelined, for example: when $c_k$ computes over $s_i$, $c_{k+1}$ can process $s_{i-1}$ and $c_{k-1}$ can process $s_{i+1}$ in parallel.

Considering that nowadays LMs operate on sequences with thousands of tokens \citep{radford2019language,brown2020language} (e.g. 2048 for GPT-3), the token dimension opens substantial space to improve the pipelining efficiency. However, applying it in practice is still challenging, especially on GPUs, for the following reasons.

\begin{figure}[t]
    \centering
    \includegraphics[ width=.35\textwidth]{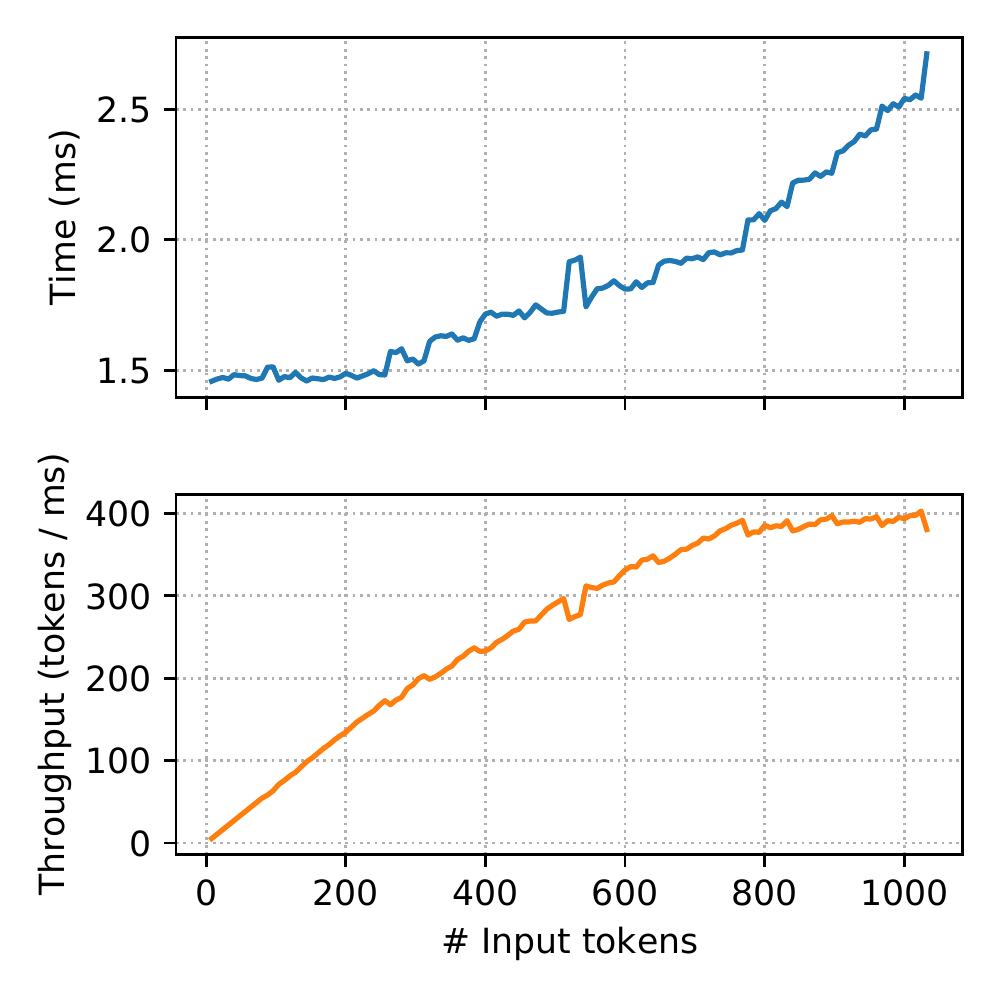}
    \vspace{-6mm}
    \caption{Forward propagation time and throughput for a single layer of GPT3-1B model with a single input sequence with different number of input tokens on a single NVIDIA V100 GPU, averaged by 30 independent runs. \textbf{Top:} Time per forward propagation. \textbf{Bottom:} Throughput measured by number of tokens per millisecond.}
    \vspace{-4mm}
    \label{fig:gpt3-1b-gpu}
\end{figure}

First, finer-grained pipelining (i.e. picking a small $|s_i|$) is prone to underutilizing the computational power of GPUs, and thus lowering the training throughput. As shown on the top part of Figure~\ref{fig:gpt3-1b-gpu}, for a single layer of the GPT3-1B model (see Table~\ref{tbl:setting} for specs), the forward propagation time for an input sequence with a single token is the same as an input sequence with 256 tokens.
In this case, the GPU is not being fully utilized for input sequence lengths less than 256.
This means a large subsequence length is needed to achieve high throughput for a single layer (see the bottom part of Figure~\ref{fig:gpt3-1b-gpu}).
On the other hand, although GPUs have better training throughput per layer for longer sequences due to the SIMD architecture and better locality, longer input slices lead to fewer pipeline stages within a sequence, which will increase the pipeline bubble, and thus reduce the pipeline efficiency and hurt the overall training speed.

Second, splitting inputs into multiple same-size chunks for pipelining, as normally done in existing work \citep{huang2019gpipe, harlap2018pipedream}, is not the ideal way for pipelining on the token dimension. For the self-attention layer, the computation of $\text{SelfAtt}(h_1)$ only requires the hidden state $h_1$ from its previous layer, while the computation of $\text{SelfAtt}(h_L)$ takes all $h_1,\ldots, h_L$ as inputs, as shown in Figure~\ref{fig:transformer}. Therefore, the computation load on a later token position in a sequence is heavier than that of previous tokens. Since the total latency of a pipeline is determined by its slowest stage (Figure~\ref{fig:non-uniform-timeline}), an optimal slicing scheme should have a long slice in the beginning and a shorter slice in the end. We next develop methods to select the optimal slicing scheme over the token dimension.

\subsection{Selecting Optimal Slicing Scheme}
\label{sec:dp}

We propose a dynamic programming (DP) algorithm to partition the input sequence to achieve the optimal pipeline efficiency. Specifically, given a partitioned Transformer-based LM $F = c_K \circ \cdots \circ c_1$ and a training input sequence of length $L$, the goal of the algorithm is to find the \emph{slicing scheme} $l_1, \ldots, l_M$ to minimize the total forward and backward propagation latency, where $l_i = |s_i|$ is the length each sub-sequence slice $s_i$ ($l_1 + \cdots + l_M = L$).

\begin{figure}[t]
    \centering
    \includegraphics[width=.47\textwidth]{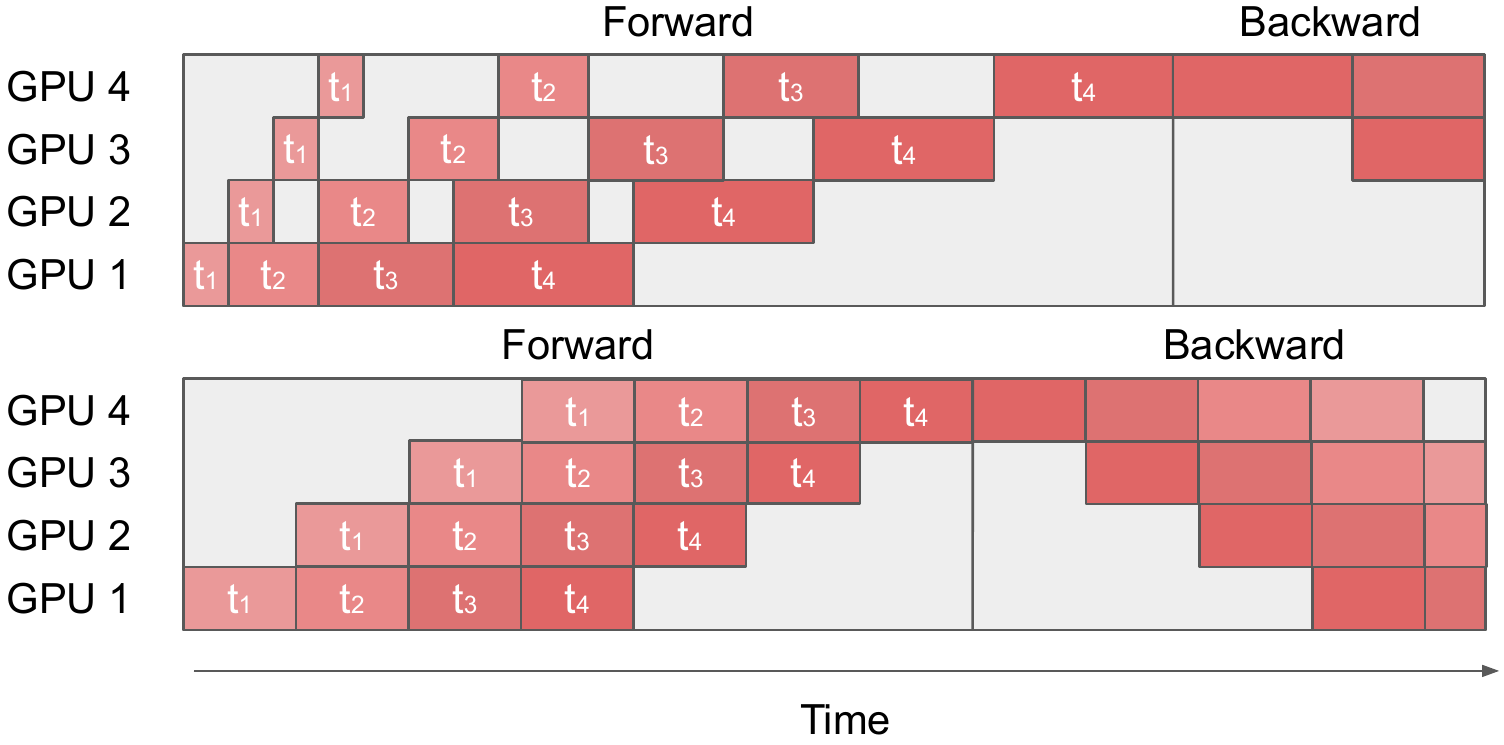}
    \caption{Execution timeline for inputs for uniform sequence split with non-uniform running time (top) and non-uniform sequence split with uniform running time (bottom). The total latency of a pipeline is determined by its slowest stage, and thus splits with non-uniform running time result in larger pipeline bubbles and inferior pipeline efficiency.}
    \vspace{-4mm}
    \label{fig:non-uniform-timeline}
\end{figure}

Let's first consider the latency of forward propagation.
As shown in Section~\ref{sec:main-technique}, all cells $c_k$ have exact same amount of computation.

The forward propagation time $t_i$ for the slice $s_i$ on the cell $c_k$ is determined by the length of the $i$th slice ($l_i$), the lengths of all the previous subsequences ($l_1,\ldots ,l_{i-1}$), and the cluster specifications (e.g., GPU, bandwidth and latency of the underlying computer networks). We use $t_{fwd}$ to denote the sum of the computation latency plus data transmission latency for a given $l_i$ and the previous subsequences $l_1,\ldots,l_{i-1}$. We have:
\begin{equation}
    t_i = t_{\mathit{fwd}}\left(l_i, \sum_{j=1}^{i-1} l_j\right).
\end{equation}
Note the second term $\sum_{j=1}^{i-1} l_j$ is the total length of previous subsequences $s_1, \ldots, s_{i-1}$ to compute $\text{SelfAtt}(s_t)$. As visualized in Figure~\ref{fig:non-uniform-timeline}, The optimal overall pipeline forward propagation latency is:
\begin{equation}
T^* = \min_{l_1, \ldots, l_M}\left\{\sum_{i=1}^M t_i + (K - 1) \cdot \max_{1 \le j \le M}\{t_j\}\right\}.
\end{equation}

The overall latency consists of two terms: The first term here is the total forward propagation time on a device (i.e. on a cell $c_k$); The second term is the overhead brought by the pipeline execution, which is determined by the slowest component in the whole pipeline multiplied by the number of pipeline stages $K$ minus 1. For example, on the top of Figure~\ref{fig:non-uniform-timeline}, the total execution time will be $T = (t_1 + \ldots + t_4) + 3t_4.$ 

Our goal is to find the optimal slicing scheme $l_1, \ldots, l_M$ that achieves the optimal latency $T^*$. We choose to first enumerate the second term $t_\mathit{max} = \max_{1 \le j \le M}\{t_j\}$ and minimize the first term for each different $t_\mathit{max}.$ In other words, we reformulate $T^*$ as:
\begin{align}
&T^* = \min_{t_\mathit{max}}\left\{S^*(L; t_\mathit{max}) + (K - 1)\cdot t_\mathit{max}\right\}, \\
&S^*(L; t_\mathit{max}) = \min_{l_1 + \cdots + l_M = L}\left\{\sum_{i=1}^M t_i \mid t_i\le t_\mathit{max} \right\}.
\end{align}

Note that $S^*(\cdot; t_\mathit{max})$ has the following optimal substructure: 
\begin{align}
    S^*(i; t_\mathit{max}) = \min_{1\le k \le i} \{ & S^*(i - k; t_\mathit{max}) + t_{\mathit{fwd}}(k, i - k) \nonumber
    \\ & \mid t_{\mathit{fwd}}(k, i - k) \le t_\mathit{max}\}.
\end{align}
Therefore, we can get the slicing scheme $l_1, \ldots, l_M$ that achieves the total total forward propagation time $S^*(L; t_\mathit{max})$ with Algorithm~\ref{alg:dp}. By enumerating all different $t_\mathit{max},$ we can get the optimal slicing scheme that reaches the optimal overall pipeline latency $T^*$.

\begin{algorithm}[tb]
  \caption{Selecting optimal slicing scheme given $t_\mathit{max}$.}
  \label{alg:dp}
\begin{algorithmic}
    \STATE {\bfseries Input:} Forward propagation time function $t_\mathit{fwd}$ and maximum per-slice time $t_\mathit{max}.$
    \STATE {\bfseries Output:} Minimal total forward propagation time $S^*(L; t_\mathit{max})$ and the corresponding slicing scheme $l_1, \ldots, l_M.$
    \STATE \emph{// Dynamic programming for the total forward propagation time.}
    \STATE $S^*(0; t_\mathit{max}) \leftarrow 0$
    \FOR{$i$ {\bfseries from} $1$ {\bfseries to} $L$}
        \STATE $S^*(i; t_\mathit{max}) \leftarrow \min_{1\le k \le i} \{S^*(i-k; t_\mathit{max}) + t_{\mathit{fwd}}(k, i - k)\mid t_{\mathit{fwd}}(k, i - k) \le t_\mathit{max}\}$.
        \STATE $q_i \leftarrow \argmin_{1\le k \le i} \{r_{i-k} + t_{\mathit{fwd}}(k, i - k)\mid t_{\mathit{fwd}}(k, i - k) \le t_\mathit{max}\}$.
    \ENDFOR
    \STATE \emph{// Derive the optimal slicing scheme.}
    \STATE $i \leftarrow L, l \leftarrow \{\}$
    \WHILE{$i > 0$}
        \STATE $l.\mathit{prepend}(q_i)$
        \STATE $i \leftarrow i - q_i$
    \ENDWHILE
\end{algorithmic}
\end{algorithm}

\noindent \textbf{Complexity.} With our DP algorithm, we can compute the best partition in $O(L^2)$ time for a fixed $t_\mathit{max}$. Note that in total there are at most $O(L^2)$ different choices ($t_\mathit{fwd}(i,j)$ for $i, j = 1, \ldots, L$) of $t_\mathit{max}$. We therefore can derive the optimal slicing scheme in $O(L^4)$ time. 

\noindent \textbf{Optimization.} To further accelerate the above DP algorithm, we enumerate different $t_\mathit{max}$ from small to large; when $K \cdot t_\mathit{max}$ is greater than the current best $T$, we stop the enumeration since larger $t_\mathit{max}$ cannot provide a better slicing scheme. In addition, during enumeration of $t_\mathit{max},$ we only evaluate with $t_\mathit{max}$ larger than the last $t_\mathit{max}$ by at least $\varepsilon$. In this case, the gap between the solution found by the DP algorithm and the global optima is at most $K\cdot \varepsilon$. We choose $\varepsilon=0.1\text{\,ms}$ in our evaluation and observe that the solution given by Algorithm~\ref{alg:dp} and the real optimal solution ($\varepsilon=0$) are always the same in all our evaluated settings. With these two optimizations, the dynamic programming can finish within a minute in our evaluations. 

\noindent \textbf{Estimating $t_\mathit{fwd}$}. To avoid the cost of evaluating $t_\mathit{fwd}(i, j)$ for all $O(L^2)$ combinations of $i, j$ on real clusters, we use a simple performance model to estimate $t_\mathit{fwd}.$ Specifically, we split  $t_\mathit{fwd}(i, j)$ into two terms:
\begin{equation}
    t_\mathit{fwd}(i, j) = t_\mathit{fwd}(i, 0) + t_\mathit{ctx}(i, j),
\end{equation}
where $t_\mathit{fwd}(i, 0)$ is the forward propagation time without any extra context input and $t_\mathit{ctx}(i, j)$ is the latency overhead brought by the extra context input. We measure the first term with all $L$ choices of $i$ and we fit a simple linear model $t_\mathit{ctx}(i, j) = a_0 + a_1i + a_2 j + a_3ij$ for the second term with a subset of all $(i,j)$ combinations. In our experiments, the linear model can achieve a $<2\%$ relative prediction error compared to the actual overhead.

The development above can be applied to backward propagation time $t_\mathit{bwd}$, since the backward propagation computation in transformers is symmetric with its forward counterpart. One step further, we can replace all the $t_\mathit{fwd}$ above with $t_\mathit{fwd} + t_\mathit{bwd}$ to derive the optimal slicing scheme that minimizes the total training time.

\begin{table*}[t]
\centering
\caption{Model settings and parallel training setups used in the evaluation. $N$: Number of Transformer layers. $H$: Hidden state size. \#Params: Number of total parameters. $L$: Input sequence length. \#GPUs: Total number of GPUs. $B$: Batch size. \#Data: Number of data parallel shards. \#Pipe: Number of pipeline stages. \#Op: Number of GPUs used for operational partitioning by each Transformer layer. }
\vspace{2mm}
\label{tbl:setting}
\scalebox{0.95}{
\begin{tabular}{rcccccccccc}
\toprule
     & Model                      & $N$                 & $H$                    & \#Params              & $L$                   & \#GPUs               & $B$ & \#Data & \#Pipe & \#Op \\ \midrule
(1)  & \multirow{3}{*}{GPT3-1B}   & \multirow{3}{*}{24} & \multirow{3}{*}{2048}  & \multirow{3}{*}{1B}   & \multirow{3}{*}{2048} & \multirow{3}{*}{192} & 128 & 8      & 24     & 1    \\
(2)  &                            &                     &                        &                       &                       &                      & 72  & 2      & 12     & 8    \\
(3)  &                            &                     &                        &                       &                       &                      & 72  & 1      & 24     & 8    \\ \midrule
(4)  & \multirow{2}{*}{GPT3-13B}  & \multirow{2}{*}{40} & \multirow{2}{*}{5120}  & \multirow{2}{*}{13B}  & \multirow{2}{*}{2048} & \multirow{2}{*}{320} & 32  & 2      & 20     & 8    \\
(5)  &                            &                     &                        &                       &                       &                      & 32  & 1      & 40     & 8    \\ \midrule
(6)  & \multirow{3}{*}{GPT3-44B}  & \multirow{3}{*}{96} & \multirow{3}{*}{6144}  & \multirow{3}{*}{44B}  & \multirow{3}{*}{2048} & \multirow{3}{*}{384} & 8   & 4      & 96     & 1    \\
(7)  &                            &                     &                        &                       &                       &                      & 8   & 2      & 24     & 8    \\
(8)  &                            &                     &                        &                       &                       &                      & 8   & 1      & 48     & 8    \\ \midrule
(9)  & \multirow{2}{*}{GPT3-175B} & \multirow{2}{*}{96} & \multirow{2}{*}{12288} & \multirow{2}{*}{175B} & \multirow{2}{*}{2048} & \multirow{2}{*}{384} & 2   & 1      & 96     & 4    \\
(10) &                            &                     &                        &                       &                       &                      & 2   & 1      & 48     & 8    \\ \bottomrule

\end{tabular}
}
\end{table*}

\begin{figure*}[t]
    \centering
    \begin{subfigure}[t]{.24\textwidth}
        \centering
        \includegraphics[width=\textwidth]{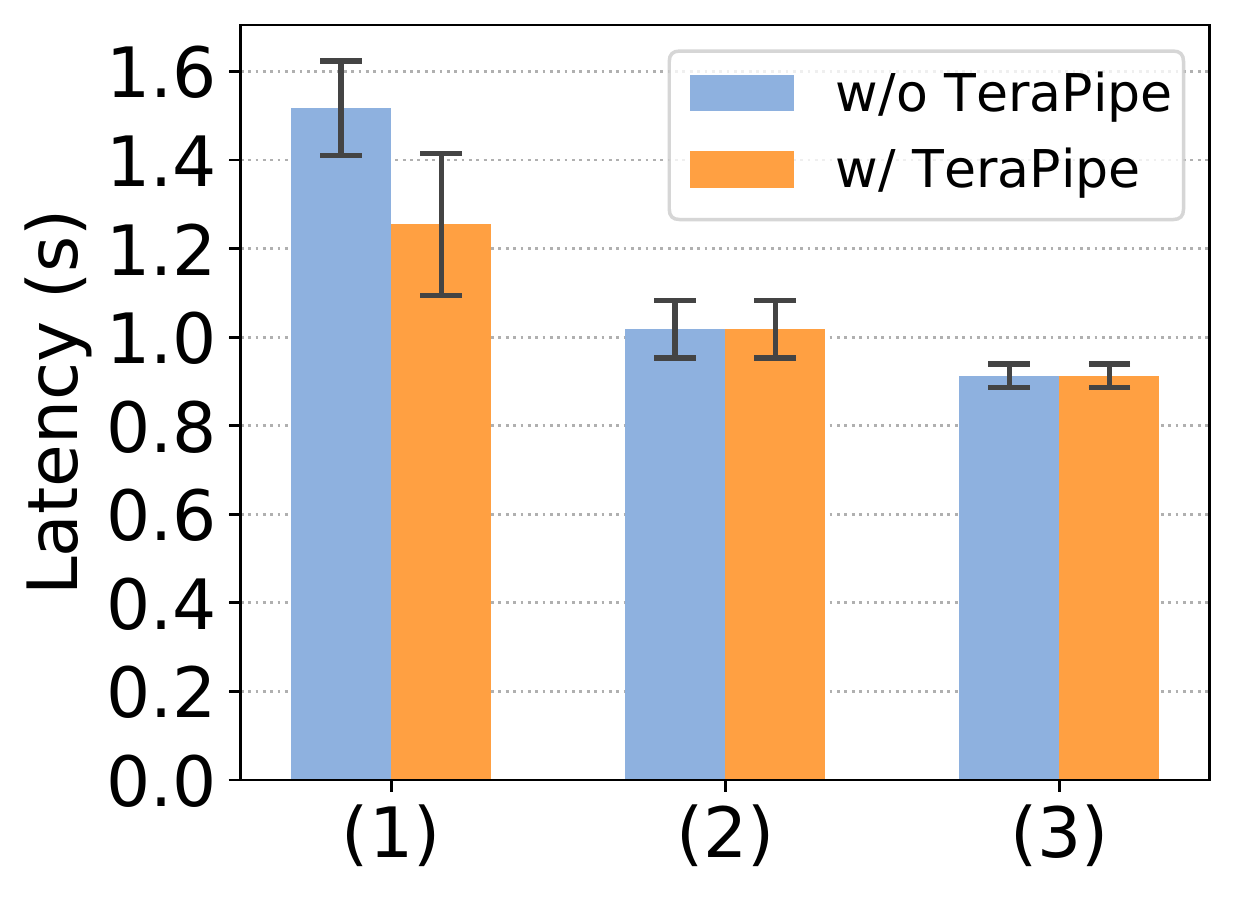}
        \caption{GPT3-1B}
        \label{fig:gpt3-1b-result}
    \end{subfigure}
    \begin{subfigure}[t]{.24\textwidth}
        \centering
        \includegraphics[width=\textwidth]{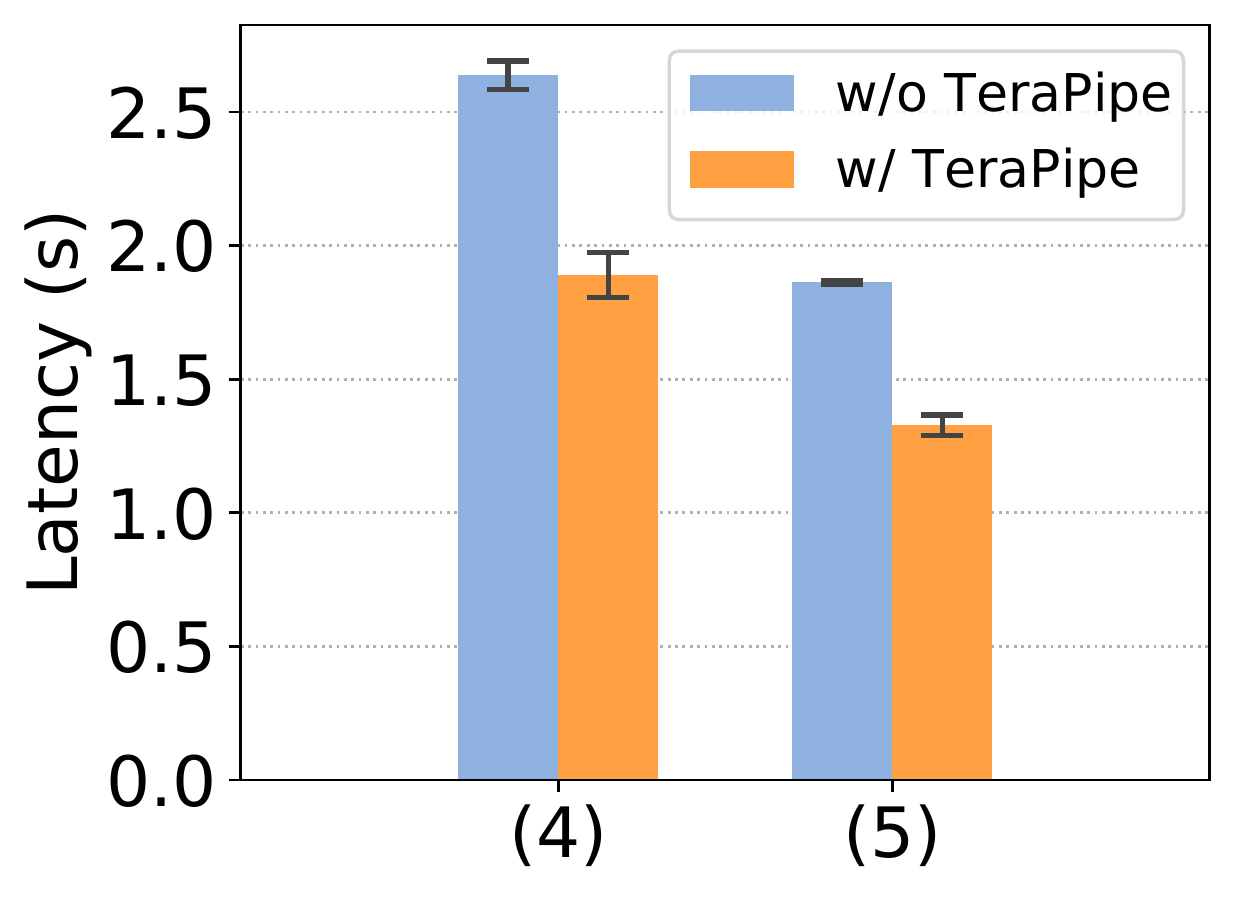}
        \caption{GPT3-13B}
        \label{fig:gpt3-13b-result}
    \end{subfigure}
    \begin{subfigure}[t]{.24\textwidth}
        \centering
        \includegraphics[width=\textwidth]{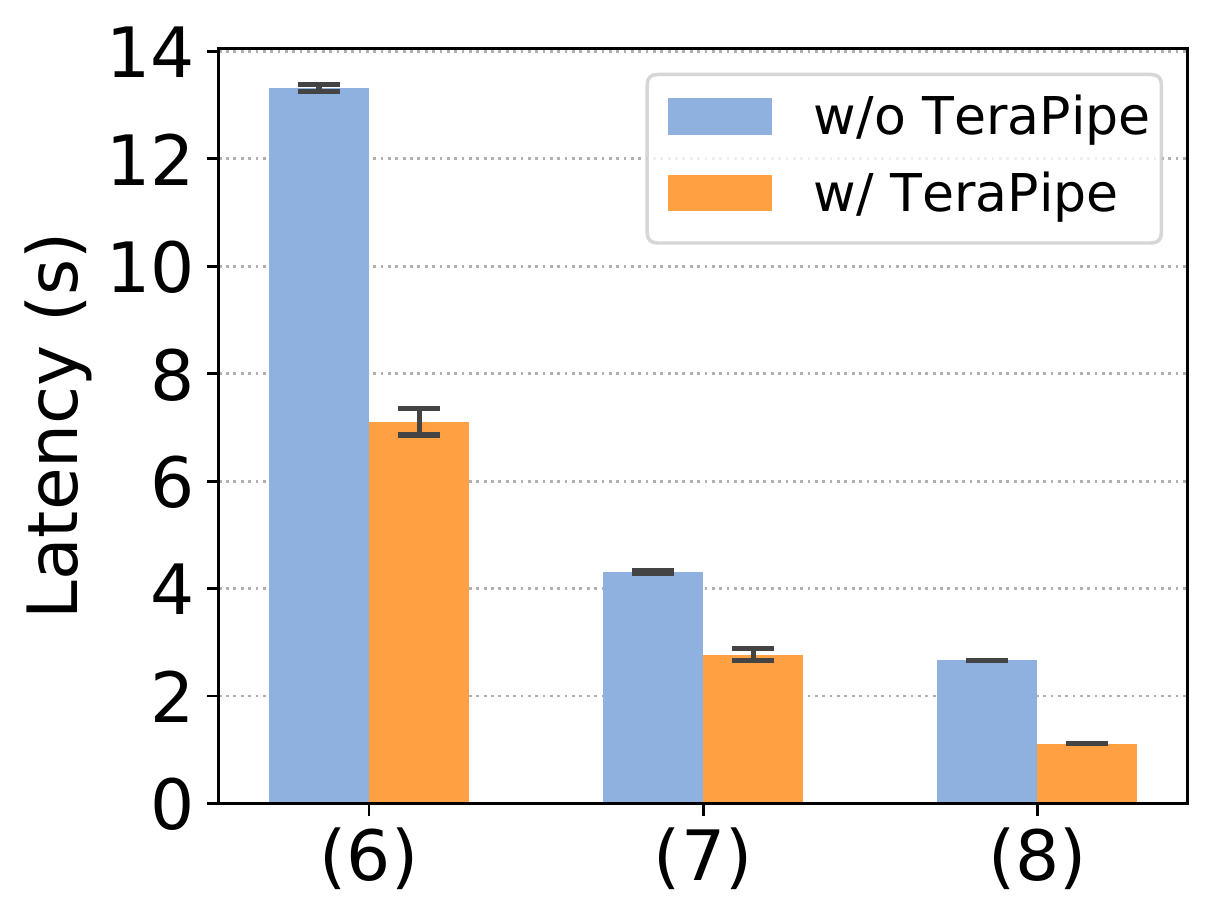}
        \caption{GPT3-44B}
        \label{fig:gpt3-44b-result}
    \end{subfigure}
    \begin{subfigure}[t]{.24\textwidth}
        \centering
        \includegraphics[width=\textwidth]{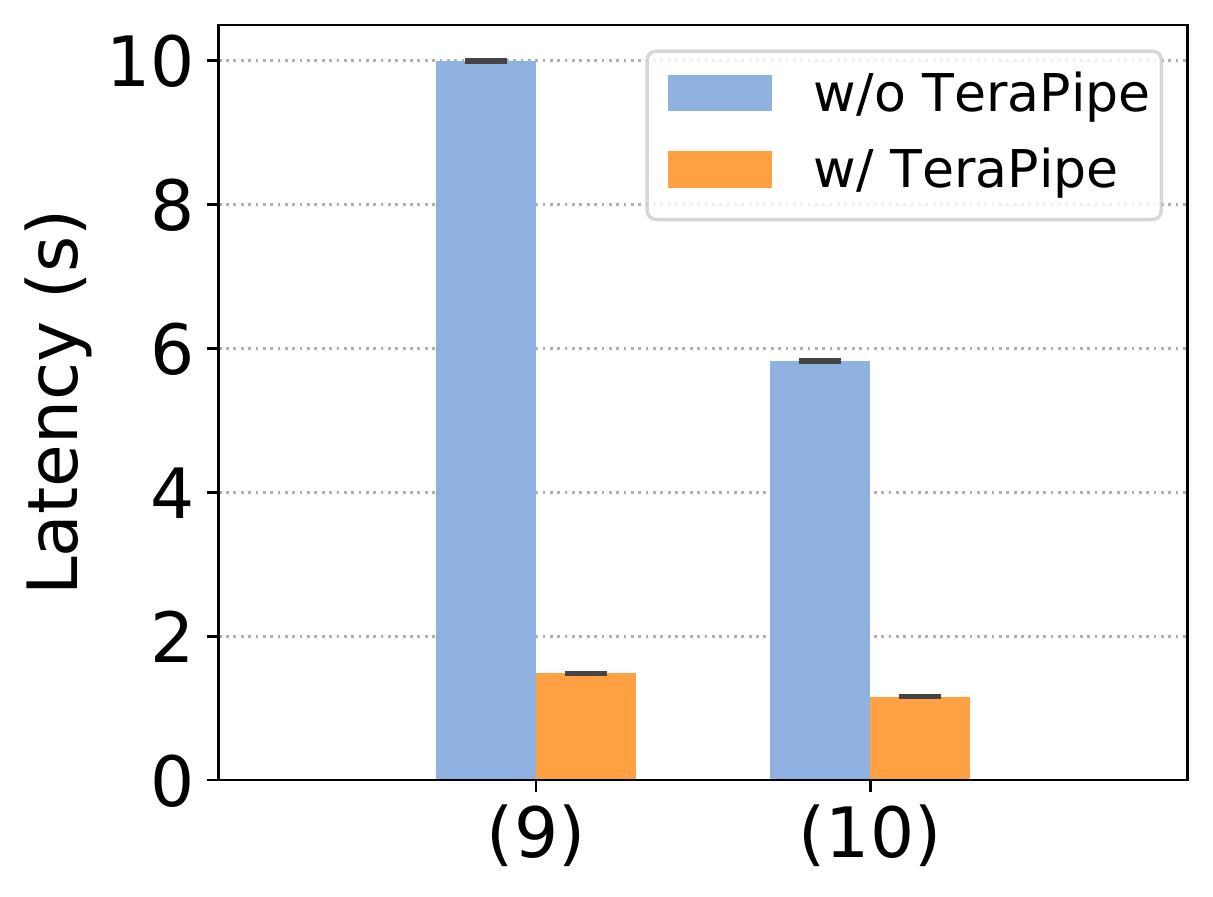}
        \caption{GPT3-175B}
        \label{fig:gpt3-175b-result}
    \end{subfigure}
        \vspace{-2mm}
    \caption{Training iteration latency for all configurations with and without \codename.  Details for each configuration are listed in Table~\ref{tbl:setting}.}
        \vspace{-2mm}
    \label{fig:main-result}
\end{figure*}

\subsection{Combining with Other Parallel Training methods}
\label{sec:combine}

The new dimension to perform pipeline parallelism proposed by \codename is orthogonal to all previous model parallel techniques, hence can be naturally combined with them. We explain next how \codename can be combined with other parallelization methods and show, when combined, it significantly boosts parallelization performance in Section~\ref{sec:evaluation}.

\noindent \textbf{Combine with microbatch-based pipeline parallelism.}
To combine with microbatch-based pipeline parallelism \citep{huang2019gpipe}, we slice the batch dimension and the token dimension jointly to form the pipeline. Specifically, consider a training input batch $(x^{(1)}, x^{(2)}, \ldots, x^{(B)}),$ where each $x^{(i)}$ is an input sequence $(x^{(i)}_1, \ldots, x^{(i)}_L)$ of length $L$, we partition the input batch into $(s^{(1)}, s^{(2)}, \ldots, s^{(D)}),$ such that each $s^{(d)}_{i}$ includes $(x^{(a)}_l, x^{(a)}_{l+1}, \ldots, x^{(a)}_{r}),$ $(x^{(a+1)}_l, x^{(a+1)}_{l+1}, \ldots, x^{(a+1)}_{r}),$ $\ldots,$ $(x^{(b)}_l, x^{(b)}_{l+1}, \ldots, x^{(b)}_{r}),$ which is the subsequence from position $l$ to $r$ of input data $a$ to $b$. During training, all slices $s^{(1)}_1, \ldots, s^{(1)}_M, s^{(2)}_1, \ldots, s^{(2)}_M, \ldots, s^{(D)}_1, \ldots, s^{(D)}_M$ can execute on cells $c_1, \ldots, c_K$ in a pipelined fashion. 
To jointly optimize the sequence slicing and batch splitting, the DP algorithm in Section~\ref{sec:dp} can be extended to include the batch dimension: we can first run the whole DP algorithm in Section~\ref{sec:dp} for all different batch sizes $b$ from $1$ to $B.$ For each $b$, we derive the optimal $T_b$ and the corresponding slicing scheme $s_b.$ With all $T_b$ and $s_b,$ we only need to determine the size of each slice in the batch dimension $b_1, \ldots, b_D$ such that $b_1 + \cdots + b_D = B$ and $T_{b_1} + \cdots + T_{b_D}$ is minimized. This reduces to a 1D knapsack problem and can be solved using off-the-shelf solvers.

\noindent \textbf{Combine with operation partitioning.}
\codename is orthogonal from operation partitioning in the sense that:
operation partitioning is \emph{intra-operation} parallelism that parallelizes the execution of a single operation, whereas \codename pipelines the execution of different operations. To combine with operation partitioning, we distribute each pipeline parallel cell $c_K$ to a set of target devices and then perform operation partitioning across target devices. 

\noindent \textbf{Combine with data parallelism.}
Similarly, because data parallelism maintains multiple identical copies of the model, we can perform model parallelism for each data parallel model replica and synchronize the gradient updates between the replicas after each forward and backward propagation. 

\noindent \textbf{Combine with memory optimization.} Same as previous pipeline parallel methods \citep{huang2019gpipe}, \codename stores the activations of a whole mini-batch in our implementation. \codename can also be combined with various memory optimization techniques, e.g., gradient accumulation \cite{fan2020dapple}, rematerialization \citep{chen2016training, jain2019checkmate}, or memory swapping \citep{ren2021zero}. See supplementary material for more discussions on combining \codename with gradient accumulation.

\section{Evaluation}
\label{sec:evaluation}
\codename is a synchronous model parallel training method that performs exactly the same underlying optimization algorithm as training the model on a single device. The optimization performance of \codename (i.e. training loss versus training iterations) is hence the same compared to training on a single device. Therefore, in this paper, we focus on the per-iteration latency (i.e. wall-clock time used per training iteration) as our evaluation metric.

We evaluate \codename following the setup in \citet{brown2020language}. Specifically, we test 3 settings in \citet{brown2020language}: GPT3-1B, GPT3-13B, and GPT3-175B, which have 1 billion, 13 billion, and 175 billion parameters in total, respectively. Note that GPT3-175B is the largest setting in \citet{brown2020language}. In addition, we also test on a GPT3-44B model with half the hidden state size $H$ of the GPT3-175B model, which includes 44 billion parameters in total.

For each model, we select multiple data parallelism, operation partitioning, and pipeline parallelism setup combinations. The configuration details are shown in Table~\ref{tbl:setting}. For all configurations, we set the input sequence length $L = 2048$ following \citet{brown2020language}. We evaluate the configurations on an AWS cluster with p3.16xlarge nodes (each with 8 NVIDIA V100 GPUs). For each model, we select a cluster size based on its model size and number of layers so that each pipeline stage (each cell $c_k$) has the same number of layers. Since operation partitioning requires higher inter-connection speed compared to pipeline parallelism, we perform operation partitioning only inside a node, where all GPUs have high-speed inter-connection thanks to NVLink. For each configuration, we select the maximal batch size that can fit the memory of the GPUs. 

We compare the per-iteration latency achieved by previous model parallel methods without \codename and the latency achieved by \codename for each configuration. Specifically, for the setup without \codename, we measure the training latency with GPipe \citep{huang2019gpipe} as the pipeline parallel training method. For \codename, we perform a joint dynamic programming on both batch and token dimension as shown in Section~\ref{sec:combine} and measure the training latency with the optimal slicing scheme found by the dynamic programming algorithm. All the latency results in the paper are averaged over 10 runs. The detailed numbers of the latency results and the solution find by the dynamic programming algorithm can be found in the supplementary material.

\begin{figure}[t]
    \centering
    \begin{subfigure}[t]{.15\textwidth}
        \centering
        \includegraphics[height=3.5cm]{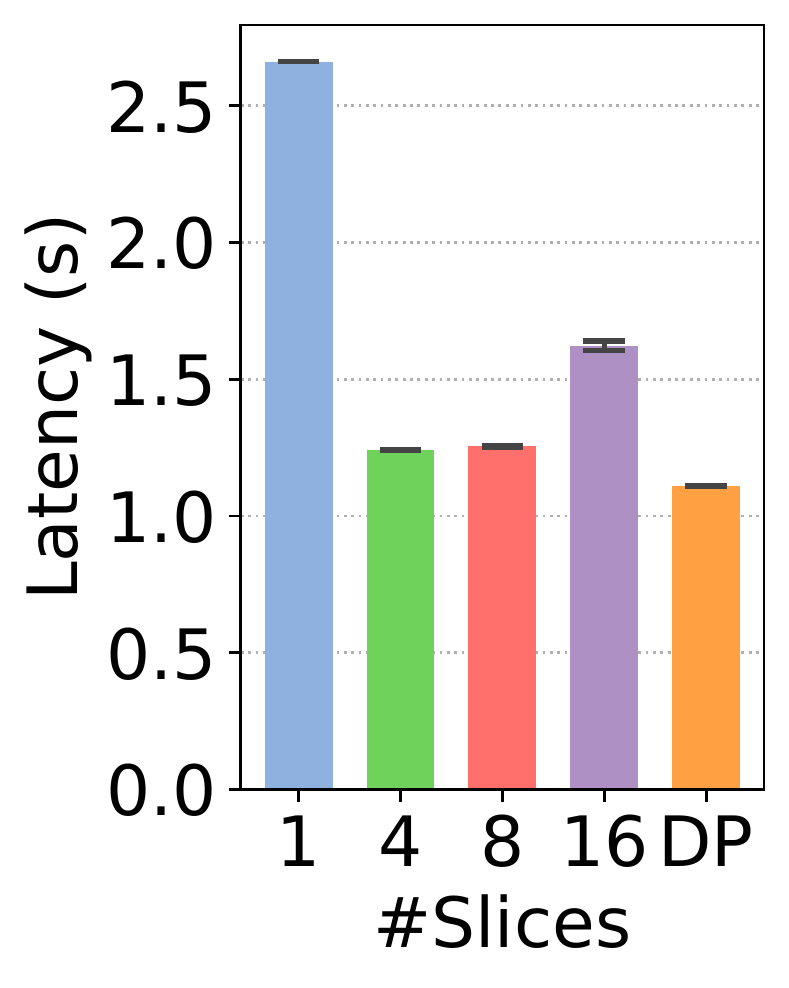}
        \caption{GPT3-44B (8)}
        \label{fig:gpt3-44b-dp}
    \end{subfigure}
    \begin{subfigure}[t]{.30\textwidth}
        \centering
        \includegraphics[height=3.5cm]{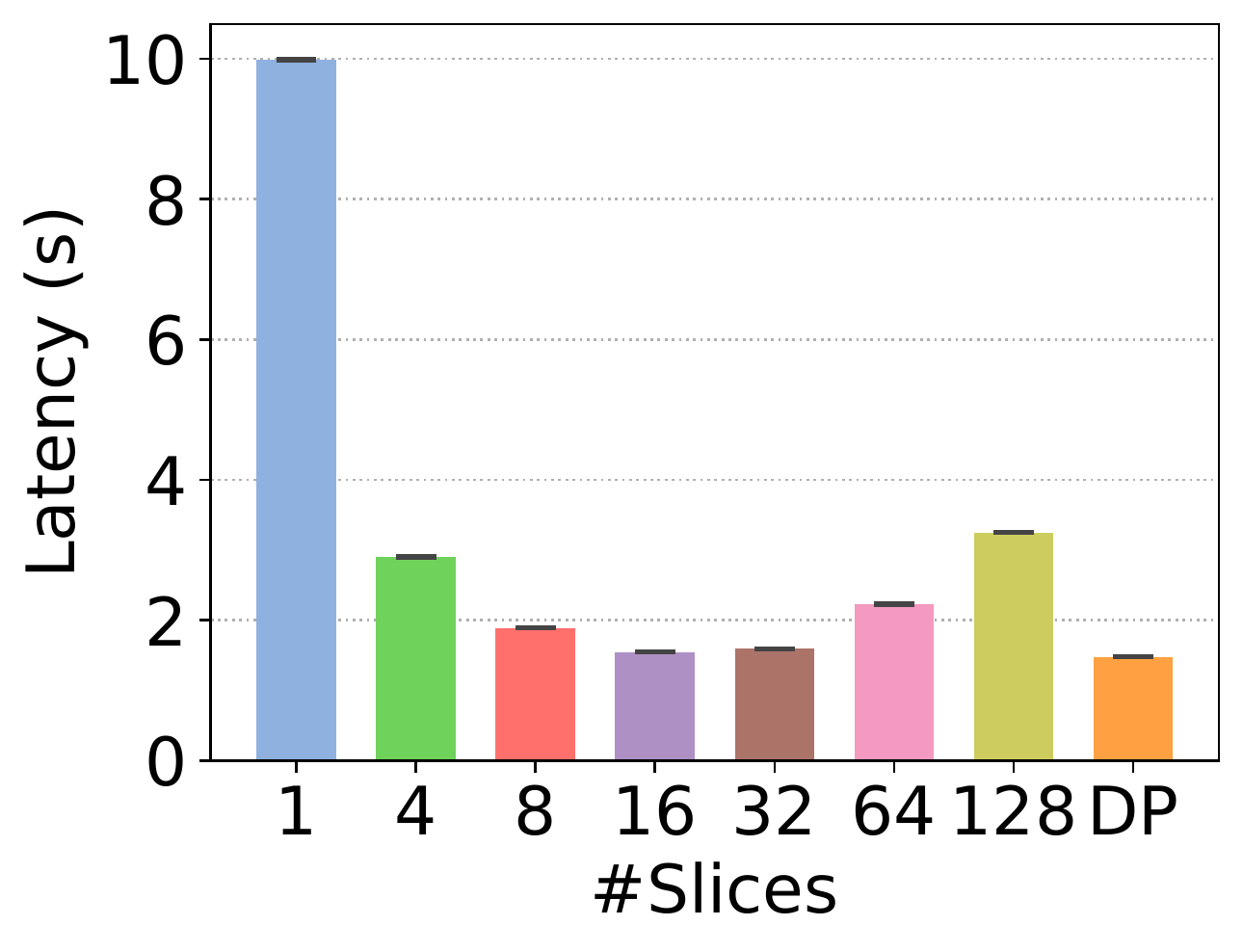}
        \caption{GPT3-175B (9)}
        \label{fig:gpt3-175b-dp}
    \end{subfigure}
    \vspace{-1mm}
    \caption{Training iteration latency of \codename with uniform slicing scheme with different number of slices and the optimal slicing scheme find by the dynamic programming algorithm.}
    \vspace{-4mm}
    \label{fig:dp-result}
\end{figure}

\subsection{Main Results}

We show the latency results for all configurations in Figure~\ref{fig:main-result}. \codename accelerates the training for all models: For GPT3-1B, \codename accelerates training for setting (1) by 1.21x. For setting (2) and (3), because of the large batch size, the optimal slicing scheme found by our dynamic programming algorithm only slices the batch dimension and thus \codename does not provide speedup. For GPT3-13B, \codename speeds up the training by 1.40x for both setting (4) and (5). For GPT3-44B, \codename accelerates the training by 1.88x, 1.56x, and 2.40x for setting (6), (7), and (8), respectively. For GPT3-175B, \codename accelerates the training by 6.75x and 5.02x for setting (9) and (10), respectively. 

\codename provides higher speedup for larger models: Larger models have a larger hidden state size $H,$ and a larger portion of GPU memory is devoted to storing the model weights and hidden states. Therefore, the batch size $B$ has to be decreased to fit the model into the GPU memory, as shown in the setup in Table~\ref{tbl:setting}. Smaller batch size $B$ limits the previous microbatch-based pipeline parallel methods' ability to saturate the pipeline bubbles, while the token dimension used by \codename still provides abundant opportunity to improve pipeline efficiency. In addition, larger models have more pipeline stages compared to smaller models, because larger models have more layers and each layer takes more memory than the smaller models. More pipeline stages require more input slices to saturate the pipeline.

\subsection{Dynamic Programming}

In this subsection, we provide an ablation study on the effectiveness of the dynamic programming algorithm proposed in Section~\ref{sec:dp}. We compare the training latency with the slicing scheme found by the dynamic programming algorithm, to a simple heuristic that slices the input sequence uniformly. Specifically, we evaluate GPT3-44B with setting (8) and GPT3-175B with setting (9). For the uniform slicing baseline, we slice the whole input on the batch dimension and range the number of slices on the token dimension from 1 to 16 and 1 to 128 for two settings, respectively, and evaluate the iteration latency for each uniform slicing scheme.

The result is shown in Figure~\ref{fig:dp-result}. As in Section~\ref{sec:main-technique}, too fine-grained pipeline (e.g. \#slices=128 in Figure~\ref{fig:gpt3-175b-dp}) performs badly because of the underutilization of the GPUs. Also, too coarse-grained pipeline (e.g. \#slices=4 in Figure~\ref{fig:gpt3-175b-dp}) has large pipeline bubbles, which leads to high iteration latency. In addition, because of the non-uniform running time brought by the Transformer structure, the slicing scheme derived by the dynamic programming program achieves better performance compared to the best uniform sliced pipeline: the optimal solutions found by dynamic programming are 1.12x and 1.04x faster compared to the best uniform slicing scheme for GPT3-44B and GPT3-175B model, respectively.

\begin{figure}[t]
    \centering
    \includegraphics[width=.33\textwidth]{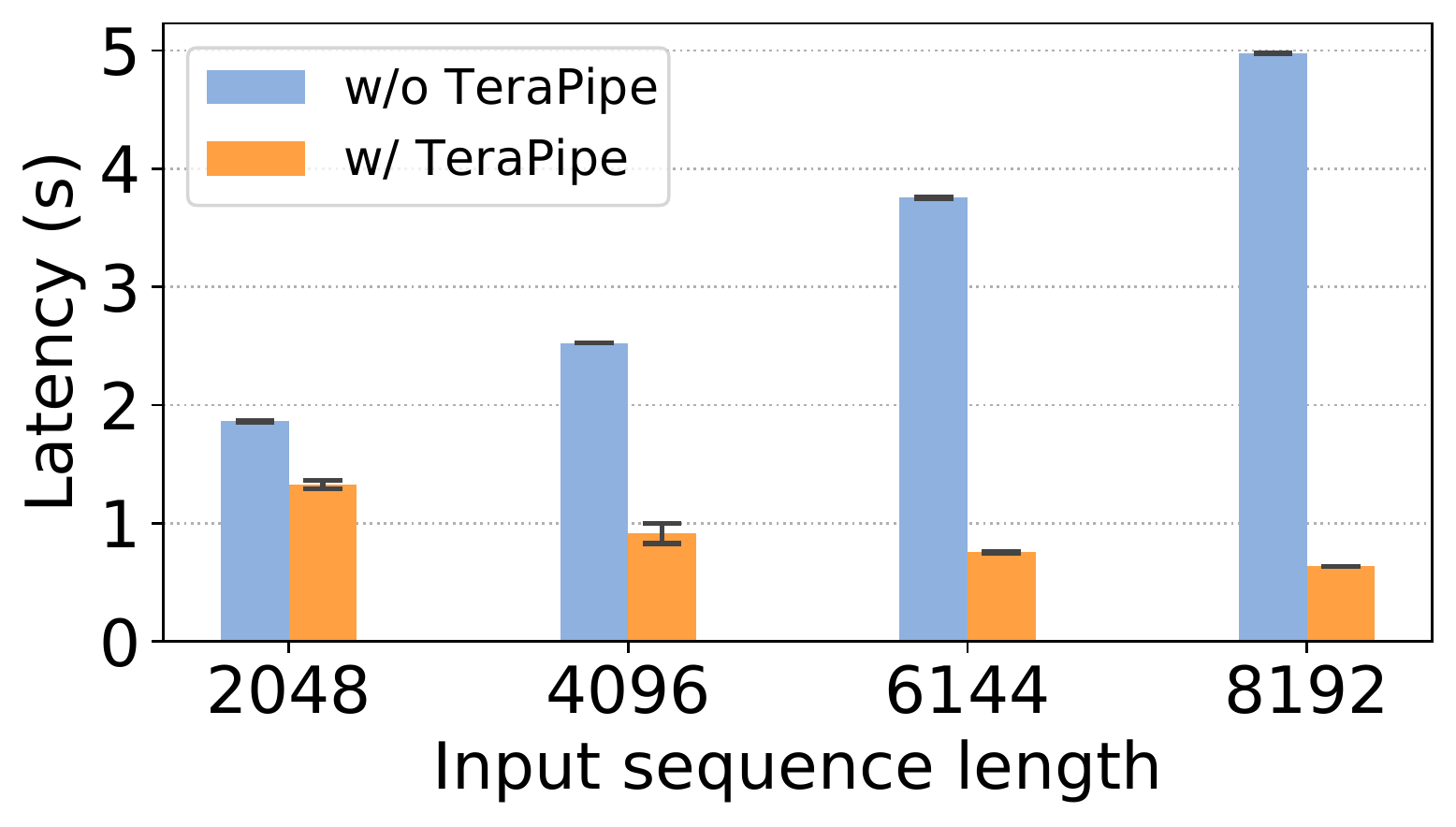}
     \vspace{-4mm}
    \caption{Training iteration latency of \codename with different input sequence length for the GPT3-13B model.}
    \vspace{-5mm}
    \label{fig:var-seqlen}
\end{figure}

\subsection{Longer Sequence Length}
A growing set of works start to focus on increasing the input sequence length of the Transformers \citep{tay2020long, zaheer2020big, kitaev2020reformer}. Long sequence length enables Transformers to reason about long-term dependencies and thus extends its applicability to more complex applications such as modeling documents. However, longer sequences increases the memory usage of a single input sequence, and decreases the maximum batch size allowed, which limits the pipeline efficiency of previous microbatch-based pipeline parallelism methods.

In this subsection, we vary the sequence length from 2048 to 8192 for the GPT3-13B model (setting (5)) and evaluate the training iteration latency. Because of the growth in memory usage, the batch sizes for sequence length 4096, 6144, 8196 are reduced to 8, 4, 2, respectively. We show the results in Figure~\ref{fig:var-seqlen}. \codename achieves 2.76x, 4.97x, 7.83x speedup for sequence length 4096, 6144, and 8196, respectively. As the sequence length grows, the gap between the performance with and without \codename significantly increases, as expected. Meanwhile, longer sequence length provides more space on the token dimension and thus \codename can perform even better -- \codename enables efficient training of future-emerging LMs with growing sequence lengths.

\section{Conclusion}
\label{sec:conclusion}

We present \codename, a high-performance token-level pipeline parallel algorithm for training large-scale Transformer-based language model. We develop a novel dynamic programming-based algorithm to calculate the optimal pipelining execution scheme, given a specific LM and a cluster configuration.
\codename is orthogonal to other model parallel training methods and can be complemented by them.
Our evaluations show that \codename accelerates the synchronous training of the largest GPT-3 models with 175 billion parameters by 5.0x on an AWS cluster with 48 p3.16xlarge instances compared to previous methods.

\section*{Acknowledgement}

We thank our anonymous reviewers for their insightful feedback. We also thank Lianmin Zheng and many others at the UC Berkeley RISELab for their helpful discussion and comments. In addition to NSF CISE Expeditions Award CCF-1730628, this research is supported by gifts from Alibaba Group, Amazon Web Services, Ant Group, CapitalOne, Ericsson, Facebook, Futurewei, Google, Intel, Microsoft, Nvidia, Scotiabank, Splunk, and VMware.

\bibliography{terapipe}
\bibliographystyle{icml2021}

\onecolumn

\appendix

\section*{Appendix}

\section{Combine \codename with Gradient Accumulation}

\codename and gradient accumulation (GA) are orthogonal and \codename can further speed up over GA. To see this, we visualize a 3-stage pipeline training with an input batch of 6 training sequences below, similar to Figure 2 in the main paper. 

\begin{center}
    \includegraphics[width=.85\textwidth]{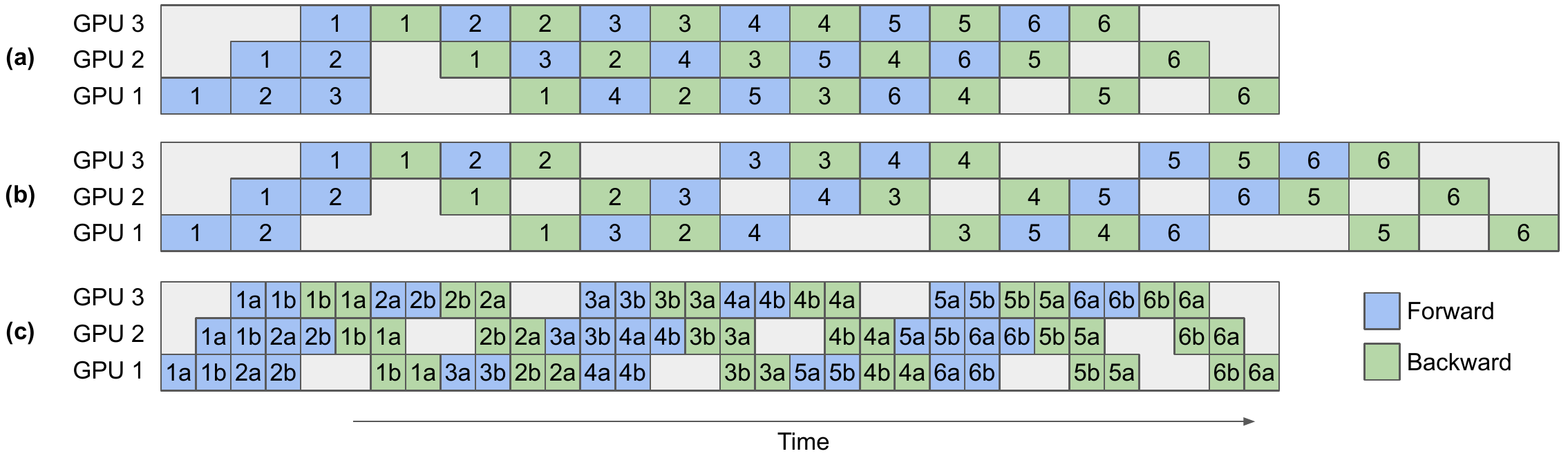}
    \vspace{-3mm}
\end{center}

In (a), we show the case where each GPU is capable of storing the intermediate activations of at most 3 input sequences. With scheduling algorithms like DAPPLE \citep{fan2020dapple}, GA indeed increases the pipeline efficiency. However in (b), when each GPU can only support 2 input sequences (due to large model size), the forward pass of input sequence 3 cannot start on GPU 1 until sequence 1 finishes the backward pass and release the memory of its intermediate activations. The memory constraint limits the pipeline efficiency: only two GPUs can work at a time, and GA cannot solve the issue. In (c), we follow the setting in (b) but enable \codename to split a training sequence into two. \codename improves the pipeline efficiency compared to (b) thanks to more fine-grained pipelining: the three can work at the same time. 

In our experiments, we have 48 pipeline stages but a single GPU is only capable to hold 2 input sequences due to its memory capacity. Even with newer GPUs (e.g. 80GB A100, 5x memory compared to V100s in the paper), their memory capacity is still not enough to fulfill the pipeline with 48 input sequences. Therefore, even with GA, \codename is still expected to significantly improve the training efficiency.

\section{Implementation}

We implement \codename with PyTorch \citep{paszke2019pytorch} and NCCL \citep{nccl}. We use Megatron-LM \citep{shoeybi2019megatron} as the library for operation partitioning and implement microbatch-based pipeline parallelism and data parallelism by ourselves. The core of \codename is implemented using 1714 lines of Python. We include the code in the supplementary material and the code will be open-sourced.

\section{Experiment Results}

Here, we include the detailed numbers (mean and standard deviation of the latency) and the slicing schemes found by the DP algorithms for all experiments in the main paper. Specifically, we list the details of Figure~\ref{fig:main-result}, \ref{fig:dp-result}, and \ref{fig:var-seqlen} in Table~\ref{tbl:detailed_evaluation_data}, \ref{tbl:detailed_dp_result_data}, and \ref{tbl:detailed_seqlen_result_data}.

\begin{table}[h]
\centering
\caption{Detailed numbers and slicing schemes in main experiments (Figure~\ref{fig:main-result} in the main paper).}
\vspace{2mm}
\label{tbl:detailed_evaluation_data}
\scalebox{0.90}{
\begin{tabular}{@{}cccccc@{}}
\toprule 
Model & Setting & Algorithm & Slicing Scheme & Latency (s) & TFlops (per GPU) \\
\midrule
\multirow{6}{*}{GPT3-1B}
& \multirow{2}{*}{\ref{fig:main-result}, (1)} &
 w/o \codename & [(1, [2048])] * 16 & $1.517\pm 0.107$ & 0.8841\\
&&
 w/ \codename & [(1, [776, 640 ,632])] * 16 & $1.254\pm0.160$ & 1.0695\\
& \multirow{2}{*}{\ref{fig:main-result}, (2)} &
 w/o \codename & [(1, [2048])] * 36 & $1.018\pm0.065$ & 2.9643 \\
&&
 w/ \codename & [(1, [2048])] * 36 & $1.018\pm0.065$ & 2.9643 \\
& \multirow{2}{*}{\ref{fig:main-result}, (3)} &
 w/o \codename & [(1, [2048])] * 72 & $0.913\pm0.027$ & 6.6105 \\
&&
 w/ \codename & [(1, [2048])] * 72 & $0.913\pm0.027$ & 6.6105 \\

\midrule

\multirow{4}{*}{GPT3-13B} &
\multirow{2}{*}{\ref{fig:main-result}, (4)}  &
 w/o \codename & [(1, [2048])] * 16 & $2.637\pm0.055$ & 3.0305 \\
&&
 w/ \codename & [(1, [1024, 1024])] * 16 & $1.891\pm0.084$ & 4.2261 \\
&
\multirow{2}{*}{\ref{fig:main-result}, (5)}  &
 w/o \codename & [(1, [2048])] * 32 & $1.863\pm0.007$ & 8.5792 \\
&&
 w/ \codename & [(1, [704, 688, 656])] * 32 & $1.328\pm0.037$ & 12.0354\\

\midrule

\multirow{6}{*}{GPT3-44B} &
\multirow{2}{*}{\ref{fig:main-result}, (6)}  &
 w/o \codename & [(1, [2048])] * 2 & $13.319\pm0.067$ & 0.2148\\
&&
 w/ \codename & [(1, [64] * 26 + [56] * 6 + [48])] * 2 & $7.103\pm0.243$ & 0.4028 \\
&
\multirow{2}{*}{\ref{fig:main-result}, (7)}  &
 w/o \codename & [(1, [2048])] * 4 & $4.311\pm0.032$ & 1.3274 \\
&&
 w/ \codename & [(1, [368, 384, 384, 368, 256, 288])] * 4 & $2.771\pm0.112$ & 2.0652 \\
&
\multirow{2}{*}{\ref{fig:main-result}, (8)}  &
 w/o \codename & [(1, [2048])] * 8 & $2.662\pm0.001$ & 4.2995 \\
&&
 w/ \codename & [(1, [384, 384, 368, 320, 296, 296])] * 8 & $1.111\pm0.002$ & 10.3018 \\

\midrule

\multirow{4}{*}{GPT3-175B} &
\multirow{2}{*}{\ref{fig:main-result}, (9)}  &
 w/o \codename & [(1, [2048])] * 2 & $9.990\pm0.005$ & 1.1300 \\
&&
 w/ \codename & [(1, [120] * 4 + [112] * 6 + [104] * 8 + [64])] * 2 & $1.481\pm0.002$ & 7.6225 \\
&
\multirow{2}{*}{\ref{fig:main-result}, (10)}  &
 w/o \codename & [(1, [2048])] * 2 & $5.822\pm0.003$ & 1.9390 \\
&&
 w/ \codename & [(1, [128] * 16)] * 2 & $1.160\pm0.001$ & 9.7318 \\

\bottomrule
\end{tabular}
}
\end{table}

\begin{table}[h]
\centering
\caption{Detailed numbers and slicing schemes in ablation studies on the effectiveness of the dynamic programming algorithm (Figure~\ref{fig:dp-result} in the main paper).}
\vspace{2mm}
\label{tbl:detailed_dp_result_data}
\scalebox{0.90}{
\begin{tabular}{@{}cccccc@{}}
\toprule 
 Model & Setting & Algorithm & Slicing Scheme & Latency (s) & TFlops (per GPU) \\
\midrule

\multirow{5}{*}{GPT3-44B} &
\multirow{5}{*}{\ref{fig:dp-result}, (a)}  &
 \#Slices=1 & [(1, [2048])] * 8 & $2.662\pm0.001$ & 4.2995 \\
 &&
 \#Slices=4 & [(1, [512] * 4)] * 8 & $1.241\pm0.003$ & 9.2226 \\
 &&
 \#Slices=8 & [(1, [256] * 8)] * 8 & $1.255\pm0.004$ & 9.1197 \\
 &&
 \#Slices=16 & [(1, [128] * 16)] * 8 & $1.241\pm0.003$ & 9.2226 \\
 &&
 DP & [(1, [384, 384, 368, 320, 296, 296])] * 8 & $1.111\pm0.002$ & 10.3018 \\

\midrule

\multirow{8}{*}{GPT3-175B} &
\multirow{8}{*}{\ref{fig:dp-result}, (b)}  &
 \#Slices=1 & [(1, [2048])] * 2 & $9.990\pm0.005$ & 1.1300 \\
&&
 \#Slices=4 & [(1, [512] * 4)] * 2 & $2.902\pm0.003$ & 3.8900 \\
&&
 \#Slices=8 & [(1, [256] * 8)] * 2 & $1.892\pm0.002$ & 5.9667 \\
&&
 \#Slices=16 & [(1, [128] * 16)] * 2 & $1.547 \pm0.01$ & 7.2973 \\
&&
 \#Slices=32 & [(1, [64] * 32)] * 2 & $1.593\pm0.002$ & 7.0866 \\
&&
 \#Slices=64 & [(1, [32] * 64)] * 2 & $2.227\pm0.002$ & 5.0691 \\
&&
 \#Slices=128 & [(1, [16] * 128)] * 2 & $3.252\pm0.004$ & 3.4714 \\
&&
 DP & [(1, [120] * 4 + [112] * 6 + [104] * 8 + [64])] * 2 & $1.481\pm0.002$ & 7.6225 \\

\bottomrule
\end{tabular}
}
\end{table}

\begin{table}[h]
\centering
\caption{Detailed numbers and slicing schemes in experiments with longer sequence lengths (Figure~\ref{fig:var-seqlen} in the main paper).}
\vspace{2mm}
\label{tbl:detailed_seqlen_result_data}
\centering
\scalebox{0.80}{
\begin{tabular}{@{}cccccc@{}}
\toprule 
Model & Input Sequence Length & Algorithm & Slicing Scheme & Latency (s)  & TFlops (per GPU) \\
\midrule

\multirow{8}{*}{GPT3-13B}
& \multirow{2}{*}{2048} &
 w/o \codename & [(1, [2048])] * 32 & $1.863\pm0.007$ & 8.5792 \\
&&
 w/ \codename & [(1, [704, 688, 656])] * 32 & $1.328\pm0.037$ & 12.0354 \\
& \multirow{2}{*}{4096} &
 w/o \codename & [(1, [4096])] * 8 & $2.526\pm0.001$ & 1.5819 \\
&&
 w/ \codename & [(1, [552, 536, 528, 512, 504, 496, 488, 480])] * 8 & $0.913\pm0.085$ & 4.3765 \\

& \multirow{2}{*}{6144} &
 w/o \codename & [(1, [6144])] * 4 & $3.754\pm0.006$ & 0.5322 \\
&&
 w/ \codename & [(1, [584, 568] + [512] * 6 + [496, 488, 472, 464])] * 4 & $0.756\pm0.008$ & 2.6427 \\
 
& \multirow{2}{*}{8192} &
 w/o \codename & [(1, [8192])] * 2 & $4.978\pm0.004$ & 0.2007\\
&&
 w/ \codename & [(1, [512] * 6 + [480] * 2 + [416] * 10)] * 2 & $0.636\pm0.001$ & 1.5707\\

\bottomrule
\end{tabular}
}
\end{table}

\end{document}